\documentclass[letterpaper,twocolumn,10pt]{article}

\usepackage{hegroup}
\usepackage{latexsym}
\usepackage{multirow}
\usepackage{array}
\usepackage{colortbl}   
\usepackage{xspace}
\usepackage{algorithm}
\usepackage{algorithmic}
\usepackage{amsmath}
\usepackage{bbm}
\usepackage{amssymb}
\usepackage{cleveref}
\usepackage{microtype}
\usepackage{inconsolata}
\usepackage{graphicx}

\AddToHook{cmd/appendix/before}{
    \crefalias{section}{appendix}
}

\providecommand{\citet}[1]{\cite{#1}}
\providecommand{\citep}[1]{\cite{#1}}

\newcommand{\mgtColorCell}[2]{%
  \begingroup
  \ifdim #1pt < 0.76pt \cellcolor[HTML]{FFFDFB}\else
  \ifdim #1pt < 0.80pt \cellcolor[HTML]{FFF7F0}\else
  \ifdim #1pt < 0.84pt \cellcolor[HTML]{FEEEDD}\else
  \ifdim #1pt < 0.87pt \cellcolor[HTML]{FDE2C8}\else
  \ifdim #1pt < 0.90pt \cellcolor[HTML]{FBD4AE}\else
  \cellcolor[HTML]{F7C08A}\fi\fi\fi\fi\fi
  #2%
  \endgroup
}
\newcommand{\mypara}[1]{\noindent{\bf {#1}.}\xspace}
\newcommand{\method}{\textit{RidgeFT}\xspace}
\newcommand{\mgtScore}[1]{\mgtColorCell{#1}{#1}}
\newcommand{\mgtBest}[1]{\mgtColorCell{#1}{\textbf{#1}}}

\newcommand{\affWHU}{\ensuremath{\clubsuit}}
\newcommand{\affAnt}{\ensuremath{\diamondsuit}}
\newcommand{\affHKUST}{\ensuremath{\spadesuit}}
\newcommand{\affJITRI}{\ensuremath{\heartsuit}}

\colorlet{corrcolor}{blue!70!green}

\newcommand{\corrsym}{%
  \textcolor{corrcolor}{\ensuremath{\dagger}}%
}

\title{When New Generators Arrive: Lifelong Machine-Generated Text Attribution via Ridge Feature Transfer}
\date{}

\author{
\textbf{Zhen Sun}\textsuperscript{\affWHU\,\affAnt\,\affHKUST}
\thanks{Work done during an internship at Ant Group.} \quad
\textbf{Yifan Liao}\textsuperscript{\affHKUST} \quad
\textbf{Zhicong Huang}\textsuperscript{\affAnt\,\corrsym} \quad
\textbf{Jiaheng Wei}\textsuperscript{\affHKUST} \\
\textbf{Cheng Hong}\textsuperscript{\affAnt} \quad
\textbf{Yutao Yue}\textsuperscript{\affHKUST\,\affJITRI} \quad
\textbf{Xinlei He}\textsuperscript{\affWHU\,\corrsym} \\
\\
\textsuperscript{\affWHU}\textit{Wuhan University} \qquad
\textsuperscript{\affAnt}\textit{Ant Group} \\
\textsuperscript{\affHKUST}
\textit{The Hong Kong University of Science and Technology (Guangzhou)} \\
\textsuperscript{\affJITRI}
\textit{Institute of Deep Perception Technology, JITRI}
}

\begin{document}

\makeatletter
\begingroup

\let\oldfnsymbol\@fnsymbol
\renewcommand{\@fnsymbol}[1]{%
  \textcolor{corrcolor}{\oldfnsymbol{#1}}%
}

\maketitle

\renewcommand{\thefootnote}{\fnsymbol{footnote}}

\footnotetext[2]{%
Corresponding authors: Zhicong Huang
(\href{mailto:zhicong.hzc@antgroup.com}
{\nolinkurl{zhicong.hzc@antgroup.com}})
and Xinlei He
(\href{mailto:xinlei.he@whu.edu.cn}
{\nolinkurl{xinlei.he@whu.edu.cn}}).
}
{\renewcommand{\thefootnote}{}\footnotetext{Our code is available at \url{https://github.com/Vincent-HKUSTGZ/RidgeFT-Lifelong-MGT}.}}

\endgroup
\makeatother

\begin{abstract}
Machine-generated text (MGT) attribution aims to identify the specific generator responsible for a given text, thereby providing fine-grained evidence for model accountability and misuse investigation.
As new large language models continue to emerge, attribution models must continuously incorporate new generators while preserving their ability to recognize previously seen ones.
Prior works have shown that this lifelong MGT attribution setting is challenging, and existing methods often struggle to achieve a stable balance between adapting to new classes and retaining old ones.
To address this issue, we propose \method, a lightweight analytic update framework that does not rely on exemplar replay.
\method trains a task-aware encoder on the initial generator set, stores compact class-wise sufficient statistics when each generator class is first observed, and then freezes the encoder for replay-free closed-form updates.
It then suppresses generator-irrelevant variation through covariance calibration, improves representation capacity with fixed random features, and updates new classes through closed-form ridge regression based on class-level sufficient statistics.
Across multi-topic evaluations with varying initial generator setups, \method consistently outperforms baselines.
It achieves the best overall macro-F1 across domains, backbones, and incremental protocols, while also improving both old-class retention and new-class adaptation.
These results suggest that feature-stable analytic updates provide a simple yet effective approach to lifelong MGT attribution.
\end{abstract}

\section{Introduction}
As generative tools powered by large language models (LLMs) become increasingly widespread~\cite{oyelude2024artificial,openclaw2026docs}, users can now conveniently rely on them for text generation and polishing.
While these capabilities improve writing efficiency, they also introduce potential risks of misuse~\cite{kumarage2024survey}.
For example, users may exploit LLMs to automatically produce large volumes of papers, news articles, reviews, and other forms of text, thereby disrupting the normal order of content production and undermining the credibility of human-authored writing~\cite{DBLP:journals/coling/WuYZYCW25}.
Against this backdrop, the effective identification of machine-generated text (MGT) has become an important problem.
To address this issue, existing studies have primarily focused on two aspects of MGT identification: binary detection and source attribution~\cite{DBLP:journals/coling/WuYZYCW25,DBLP:conf/ccs/0001SC0024}.
Compared with binary detection, MGT attribution further aims to identify the specific source generator, thereby providing finer-grained evidence for accountability tracking and misuse investigation~\cite{DBLP:conf/emnlp/CavaT25}.

\begin{figure}[t]
    \centering
    \includegraphics[width=1\linewidth]{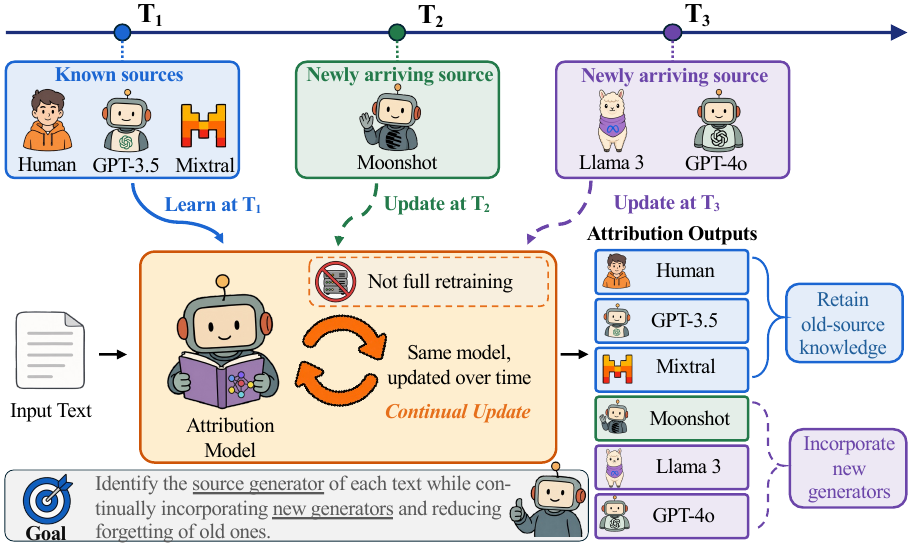}
    \caption{Illustration of the lifelong MGT attribution setting.}
    \label{fig:task_overview}
\end{figure}

Existing MGT attribution methods typically assume a fixed set of generators, which is often an unrealistic assumption in real-world settings.
Since the attribution systems operate in a dynamic and open generator space in practice, they should not only recognize newly emerging generators, but also preserve the ability to distinguish previously seen ones.
\citet{DBLP:conf/kdd/LiuZL0ZWGT00025} introduce a more realistic setting, known as class-incremental MGT attribution, which we refer to as lifelong MGT attribution throughout this paper and illustrate in \Cref{fig:task_overview}.
Under this setting, the attribution model needs to be continuously updated as new generator classes arrive over time.
However, due to factors such as computational cost, data licensing constraints, or the unavailability of historical data, it is usually impractical to recollect all past data and retrain the model from scratch~\cite{verwimp2023continual,huang2024mitigating}.
At the same time, directly updating the model using only data from new classes often leads to catastrophic forgetting~\cite{mccloskey1989catastrophic,french1999catastrophic}.
Therefore, a central challenge in lifelong MGT attribution is efficiently incorporating new generator classes under limited data while maintaining stable recognition of previously learned generators.

Under this challenge, we posit that the difficulty of lifelong MGT attribution does not necessarily need to be addressed by continuously updating the entire text encoder.
Instead, a task-tuned encoder trained on the initial set of generators can already capture strong generator-related representations.
If this encoder continues to be fine-tuned during the incremental stage, the representation space will continue to shift as new classes arrive, thereby making the decision boundaries of old classes unstable~\cite{caccia2021new,yu2020semantic}.
We therefore seek to decouple the learning of new generators from deep representation updating, ensuring that incremental knowledge is absorbed without altering or disrupting the stable representation space.
Motivated by this idea, we propose \method, an exemplar-free analytic update framework for lifelong MGT attribution. 
We consider a practical deployment scenario where the attribution system is trained and maintained from the initial stage. 
During initialization, \method uses the initial-class data to train a task-aware encoder and construct compact class-wise sufficient statistics; after that, the raw texts of old classes are discarded and never replayed. 
When new generators arrive, \method keeps the encoder frozen, maps the new data through covariance calibration and fixed random features, accumulates the corresponding statistics, and updates the classifier by a closed-form ridge solution. 
In this way, incremental learning is performed through statistical memory rather than historical-text replay or repeated encoder fine-tuning.
We evaluate \method in a multi-topic setting~\cite{DBLP:conf/kdd/LiuZL0ZWGT00025,DBLP:conf/acl/0001Z0ZL0Z025}, using P3, P4, and P5 protocols that start from 3, 4, and 5 initial classes and incrementally add 3, 2, and 1 new generator classes, respectively.
Under the standard P5 protocol, \method achieves 0.886 full-F1, 0.902 old-class F1, and 0.804 new-class F1, improving full-F1 by 0.037 over the strongest continual-learning baseline.

Our contributions are summarized as follows:
\begin{itemize}
    \item We identify lifelong MGT attribution as a generator-evolving attribution problem, where the model must preserve generator-specific decision boundaries while suppressing topic-, domain-, and length-induced nuisance variation under an exemplar-free update constraint.
    \item We propose \method, an exemplar-free analytic update framework that combines fractional covariance calibration, isotropic random feature lifting, and class-balanced closed-form ridge regression, enabling incremental updates using only compact class-wise sufficient statistics.
    \item We conduct extensive experiments across multiple topics, multiple target generators, and two backbones. The results show that \method's main advantage lies in substantially improving new-generator adaptation while maintaining competitive old-generator retention.
\end{itemize}

\section{Related Work}
Establishing effective regulatory mechanisms for MGT has become essential for maintaining content credibility and supporting platform governance~\cite{DBLP:journals/coling/WuYZYCW25}.
Recent research has pushed MGT detection from idealized binary classification toward more complex real-world scenarios. Studies have not only developed dynamic benchmarks accounting for multilingual settings and model evolution~\cite{DBLP:conf/acl/MackoKMS25,DBLP:conf/acl/YuYLC0YS25}, but also improved detector generalization to unseen generators and domains~\cite{DBLP:conf/acl/HaoLZYM25,DBLP:conf/acl/Jiao0ZG025,DBLP:conf/emnlp/ChenHHZF25}, alongside enhancing robustness against adversarial attacks~\cite{DBLP:conf/acl/LiZLSL25,DBLP:conf/naacl/LiYTJSCSS25,DBLP:conf/acl/PedrottiPCM0DE25}.
Additionally, phenomena with blurred boundaries, such as human-AI collaborative writing, have also become relevant to platform monitoring~\cite{DBLP:conf/acl/SuWWZL25,DBLP:conf/acl/SahaF25,DBLP:conf/acl/0001Z0ZL0Z025}.
However, although MGT detection techniques have made progress in addressing realistic challenges, merely determining whether a text is machine-generated is no longer sufficient to satisfy the growing demands for accountability tracing and copyright attribution.

\begin{figure*}[t]
    \centering
    \includegraphics[width=1\linewidth]{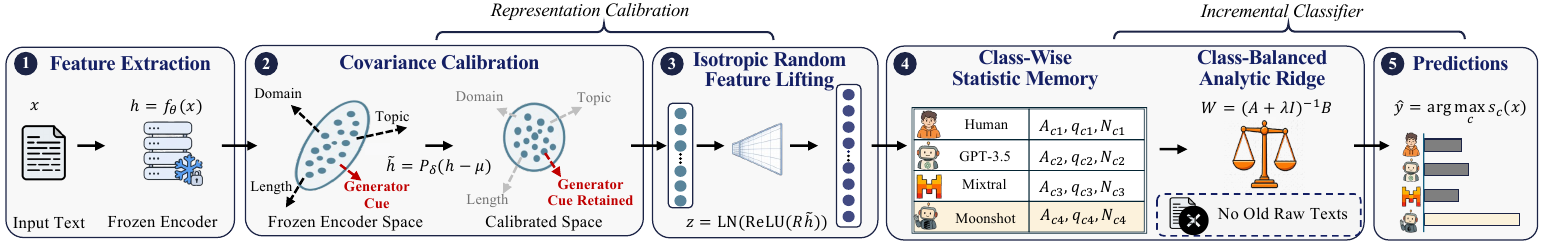}
    \caption{Overview of \method.}
    \label{fig:framework}
\end{figure*}

Compared with binary detection, MGT attribution further requires identifying the specific generator responsible for a text, providing critical evidence for model accountability and forensic analysis~\cite{DBLP:conf/ccs/0001SC0024,fang2025could}, and has emerged as an important branch of authorship attribution~\cite{huang2025authorship}.
While prior studies have explored practical attribution settings~\cite{sarvazyan2023overview,la2025authorship,najjar2025leveraging}, they mostly consider static scenarios where the candidate generator set is fixed.
In real-world deployment, rapid LLM updates require attribution models to evolve accordingly to recognize new generators.
Simultaneously, data privacy and licensing restrictions often prevent full retraining on historical data.
To address this, \citet{DBLP:conf/kdd/LiuZL0ZWGT00025} pioneered class-incremental MGT attribution and evaluated mainstream continual learning methods.
Nevertheless, existing approaches still struggle to reliably preserve recognition performance on previously seen generators while learning the characteristics of new ones.
Breaking this trade-off between adaptation to new classes and retention of old ones in lifelong MGT attribution remains an unresolved challenge, which is the central problem addressed in this work.

\section{Methodology}
In lifelong MGT attribution, continuously fine-tuning the text encoder on new generators alters the representation geometry of old classes, thereby degrading previous decision boundaries.
To address this, we propose \method, which reformulates lifelong attribution as an analytic ridge regression problem based on sufficient statistics.
\method is designed for an attribution system that is trained and maintained from the initial stage, so the required statistics of initial classes can be recorded when those classes are first available.
By freezing the base text encoder, \method preserves prior knowledge and performs analytic updates exclusively on the extracted features.
Given an input text $x$, its frozen representation $h=f_\theta(x)$ is mapped to the final prediction via a novel sequential pipeline: covariance calibration, isotropic random feature lifting, and class-balanced ridge regression ($h\rightarrow \tilde h\rightarrow z(x)\rightarrow \hat y$).
The calibration transform and the initial sufficient statistics are computed once from the base-stage training data and then stored.
During later incremental stages, \method only processes newly arriving class data and does not revisit old raw texts.

\mypara{Covariance Calibration}
Base representations may encode generator-irrelevant variations (e.g., topic, length, domain) as high-variance directions, which impairs subsequent inner-product classifiers.
To mitigate this, \method applies a mild fractional whitening transformation to suppress within-class noise while preserving the original discriminative geometry.
This transform is estimated only from the base-stage training representations and is kept fixed after initialization, so incremental updates do not require replaying previous raw texts.

First, we calculate the within-class scatter matrix $S_w$:
\begin{equation}
S_w=\frac{1}{N-C_0}\sum_{c=1}^{C_0}\sum_{i:y_i=c}(h_i-\mu_c)(h_i-\mu_c)^\top,
\end{equation}
where $N$ is the total number of base samples, $C_0$ is the number of base classes, $h_i$ is the representation of the $i$-th sample, and $\mu_c$ is the feature mean of class $c$.
To address the instability of high-dimensional covariance estimation, we apply trace-scaled shrinkage:
\begin{equation}
S_w^{\text{shrink}}=(1-\alpha)S_w+\alpha\frac{\operatorname{tr}(S_w)}{d_h}I_{d_h},
\end{equation}
where $\alpha$ is the shrinkage parameter, $\operatorname{tr}(\cdot)$ denotes the matrix trace, $d_h$ is the feature dimension, and $I_{d_h}$ is the identity matrix.
Because $d_h$ is large relative to the per-class sample size, the empirical scatter $S_w$ is poorly conditioned, and directly whitening its near-zero eigenvalues would amplify estimation noise.
Following~\citet{ledoit2004well}, we pull $S_w$ toward the scaled identity $\tfrac{\operatorname{tr}(S_w)}{d_h}I_{d_h}$, keeping $S_w^{\text{shrink}}$ well-conditioned and invertible while preserving its trace.
Finally, we compute the fractionally whitened representation $\tilde{h}$ via the eigenvalue decomposition of $S_w^{\text{shrink}}$:
\begin{equation}
S_w^{\text{shrink}}=U\Lambda U^\top,
\Lambda=\operatorname{diag}(\sigma_1,\ldots,\sigma_{d_h}).
\end{equation}

Given the eigendecomposition of \(S_w^{\mathrm{shrink}}\), we calibrate each
principal direction according to its estimated within-class variance:
\begin{equation}
P_\delta=U(\Lambda+\epsilon I_{d_h})^{-\delta}U^\top,
\tilde h=P_\delta(h-\mu).
\end{equation}

Here, $U$ and $\Lambda$ represent the eigenvectors and eigenvalues of $S_w^{\text{shrink}}$, respectively, with $\sigma_j$ denoting the $j$-th eigenvalue.
The parameter $\delta \in [0,1]$ controls the calibration strength, $\mu$ is the global feature mean, and $\epsilon$ is a small constant ensuring numerical stability.
At $\delta=0.5$ this recovers conventional whitening of the trace-shrunk within-class scatter, whereas $\delta=0$ applies no calibration and $\delta=1$ corresponds to stronger inverse-covariance scaling. A moderate exponent attenuates dominant within-class variations without excessively distorting the original feature space, providing a stable input for the downstream analytic classifier. 
As exemplified in~\Cref{fig:framework}, this calibration is intended to reduce nuisance directions associated with topic, length, and domain, while preserving generator-related cues for attribution.

\mypara{Isotropic Random Feature Lifting}
After covariance calibration, \method further lifts the calibrated representations into a fixed nonlinear random feature space to boost the expressive capacity of the analytic classifier without sacrificing closed-form incremental updates.
By adopting a non-trainable random mapping~\cite{rahimi2007random} instead of a learnable projection layer, we ensure the feature basis remains strictly invariant upon the arrival of new generators.

Let $\tilde h\in\mathbb{R}^{d_h}$ denote the calibrated encoder representation, where $d_h$ is the feature dimension.
\method samples a Gaussian random matrix $R\in\mathbb{R}^{d_\phi\times d_h}$ once before incremental learning, with each entry drawn independently as $R_{ij}\overset{\mathrm{i.i.d.}}{\sim}\mathcal{N}(0,1/d_h)$. The lifted feature is computed as
\begin{equation}
z(x)=\operatorname{LN}\left(\operatorname{ReLU}(R\tilde h)\right)\in\mathbb{R}^{d_\phi},
\end{equation}
where $d_\phi$ is the random feature dimension and $\operatorname{LN}(\cdot)$ denotes parameter-free layer normalization without learnable affine parameters.
The matrix $R$ is fixed throughout all incremental stages, so only the sufficient statistics of the ridge classifier need to be updated.

We use an isotropic Gaussian projection rather than a data-dependent projection learned from base classes.
In lifelong MGT attribution, future generators may introduce variation directions that are not present in the initial generator set.
A projection fitted to base-class geometry may therefore overemphasize existing class separations and reduce coverage of unseen generator-specific directions.
By contrast, isotropic random features provide a task-agnostic nonlinear basis that expands the calibrated representation more uniformly, enabling the downstream ridge regressor to form richer decision boundaries without changing the representation space during continual learning.

\mypara{Class-Balanced Analytic Ridge Regression}
Using the random feature representation $z(x)$, \method trains the classification head via closed-form ridge regression.
Because this analytic approach relies only on first- and second-order sufficient statistics, unlike iterative SGD-based methods, we only update the sufficient statistics for each class during the incremental stage, strictly eliminating the need to store raw text from old classes.

For each class $c$, \method stores only three sufficient statistics in the random feature space: the second-order statistic $A_c=\sum_{i:y_i=c}z_i z_i^\top$, the first-order statistic $q_c=\sum_{i:y_i=c}z_i$, and the sample count $N_c$.
These statistics are computed when class $c$ is first observed and are then retained as statistical memory, so later updates do not require replaying its raw texts.
When a new generator arrives, \method only computes the same statistics for the new class and appends them to the existing table.
However, directly summing class statistics can bias the ridge solution toward classes with larger sample counts, since both $A_c$ and $q_c$ scale with $N_c$.
To reduce this imbalance, \method introduces inverse-frequency weights based on class sample counts:
\begin{equation}
\label{eq:class_balance_weight}
\omega_c=
\frac{(N_c+\tau)^{-\beta}}
{\frac{1}{|\mathcal{Y}_t|}\sum_{c'\in\mathcal{Y}_t}(N_{c'}+\tau)^{-\beta}},
\qquad c\in\mathcal{Y}_t,
\end{equation}
where $\mathcal{Y}_t$ is the set of classes observed up to incremental stage $t$, $\beta$ controls the strength of reweighting, and $\tau$ prevents extremely small classes from receiving excessively large weights.
We then construct the ridge regression solution using the weighted statistics.
Specifically, let $\bar A=\sum_{c\in\mathcal{Y}_t}\omega_c A_c$, and let the $c$-th column of $\bar B$ be $\omega_c q_c$. The final classifier is given by
\begin{equation}
W=(\bar A+\lambda I_{d_\phi})^{-1}\bar B,
\end{equation}
where $\lambda>0$ is the ridge regularization coefficient.

By mitigating the dominance of majority classes via reweighting, \method converts incremental updates into the simple accumulation of sufficient statistics and a single linear system solve.
This enables the efficient absorption of new generators without encoder drift. At inference, the model predicts the class maximizing $s(x)=z(x)^\top W$. \Cref{alg:ridgeft_short} summarizes this overall procedure.

\begin{algorithm}[h!]
\caption{\method Update}
\label{alg:ridgeft_short}
\begin{algorithmic}[1]
\REQUIRE Frozen encoder $f_\theta$, calibration transform $P_\delta$, random matrix $R$, streams $\{\mathcal{D}_t\}_{t=0}^{T}$
\ENSURE Ridge classifier $W$

\STATE Initialize seen classes $\mathcal{Y}_{-1}\leftarrow\emptyset$ and statistics $\{A_c,q_c,N_c\}$ as empty.
\FOR{$t=0,\ldots,T$}
    \STATE $\mathcal{Y}_t \leftarrow \mathcal{Y}_{t-1}$.
    \FOR{each $(x_i,y_i)\in\mathcal{D}_t$}
        \IF{$y_i\notin\mathcal{Y}_t$}
            \STATE Add $y_i$ to $\mathcal{Y}_t$ and initialize $A_{y_i},q_{y_i},N_{y_i}$ as zero.
        \ENDIF
        \STATE $z_i \leftarrow \operatorname{LN}\!\left(\operatorname{ReLU}\!\left(RP_\delta(f_\theta(x_i)-\mu)\right)\right)$.
        \STATE $A_{y_i}\leftarrow A_{y_i}+z_i z_i^\top,\quad
        q_{y_i}\leftarrow q_{y_i}+z_i,\quad
        N_{y_i}\leftarrow N_{y_i}+1$.
    \ENDFOR
    \STATE $\displaystyle \omega_c \leftarrow
    \frac{(N_c+\tau)^{-\beta}}
    {|\mathcal{Y}_t|^{-1}\sum_{c'\in\mathcal{Y}_t}(N_{c'}+\tau)^{-\beta}},
    \quad \forall c\in\mathcal{Y}_t$.
    \STATE $\displaystyle \bar A\leftarrow \sum_{c\in\mathcal{Y}_t}\omega_c A_c,\quad
    \bar B\leftarrow [\omega_c q_c]_{c\in\mathcal{Y}_t}$.
    \STATE $\displaystyle W\leftarrow(\bar A+\lambda I_{d_\phi})^{-1}\bar B$.
\ENDFOR

\STATE \textbf{return} $W$
\end{algorithmic}
\end{algorithm}

\section{Experimental Setting}
\mypara{Task Protocol}
To simulate the continual emergence of generators in real-world scenarios, we adopt a class-incremental MGT attribution benchmark targeting $1$ Human source and $5$ LLMs.
We evaluate our method under $3$ incremental protocols: P3, P4, and P5, which begin with 3, 4, and 5 initial classes, respectively.
While P5 follows the standard single-step incremental setting of prior work~\cite{DBLP:conf/kdd/LiuZL0ZWGT00025}, P3 and P4 are introduced to further increase task difficulty.
By starting with smaller base sets and introducing one new class per subsequent step, P3 and P4 involve more incremental stages, thereby imposing substantially stricter requirements on the model's resistance to catastrophic forgetting.

\mypara{Dataset}
We conduct experiments on two representative benchmark datasets, which correspond to two major real-world scenarios where LLM misuse poses particularly significant risks: rigorous academic writing and diverse social media interaction.
(1) MGT-Academic~\cite{DBLP:conf/kdd/LiuZL0ZWGT00025}: This dataset focuses on the academic writing domain and covers three disciplines, namely STEM, Humanities, and Social Science.
It includes human-authored text as well as text generated by five mainstream models, namely GPT-3.5\footnote{https://openai.com/index/chatgpt.}, GPT-4o-mini~\cite{hurst2024gpt}, Moonshot~\cite{moonshot2026official}, Mixtral-8x7B~\cite{jiang2024mixtral}, and Llama-3.1~\cite{grattafiori2024llama}.
We use the complete dataset released by the original authors in our experiments.
(2) AIGTBench (Social Media Subset)~\cite{DBLP:conf/acl/0001Z0ZL0Z025}: The texts in this subset are drawn from major social media platforms.
It likewise contains human-written text and text generated by GPT-3.5, GPT-4o-mini, Llama-1~\cite{touvron2023llama}, Llama-2~\cite{touvron2023llama2}, and Llama-3.1. To ensure rigorous evaluation and class balance, we uniformly sample $15K$ instances from each class for our experiments.

\begin{table*}[t]
    \centering
    \caption{Performance Comparison across Domains and Models. $\ast$ denotes methods using data replay. ``Ori.'' denotes the original base-stage model before incremental updating, i.e., the S0 performance on the initial generator classes, and is reported only as a reference rather than as a continual-learning baseline. Darker color indicates better performance. Bold values indicate the best result among continual-learning methods within each backbone block.}
    \label{tab:P5_full_f1}
    \resizebox{\textwidth}{!}{
    \begin{tabular}{llcccccccccccccccc}
    \toprule
    \multirow{3}{*}{\textbf{Domain}} & \multirow{3}{*}{\textbf{New Model}} & \multicolumn{8}{c}{\textbf{RoBERTa-base}} & \multicolumn{8}{c}{\textbf{DeBERTa-base}} \\
    \cmidrule(lr){3-10} \cmidrule(lr){11-18}
    & & Ori. & LwF$\ast$ & iCaRL$\ast$ & BiC$\ast$ & EASE$\ast$ & PASS & SimpleCIL & \method (Ours) & Ori. & LwF$\ast$ & iCaRL$\ast$ & BiC$\ast$ & EASE$\ast$ & PASS & SimpleCIL & \method (Ours) \\
    \midrule
    
    \multirow{6}{*}{Social Science}
    & GPT-3.5     
    & 0.906 & \mgtScore{0.833} & \mgtScore{0.845} & \mgtScore{0.835} & \mgtScore{0.844} & \mgtScore{0.772} & \mgtScore{0.769} & \mgtBest{0.865}
    & 0.891 & \mgtScore{0.819} & \mgtScore{0.812} & \mgtScore{0.801} & \mgtScore{0.753} & \mgtScore{0.780} & \mgtScore{0.765} & \mgtBest{0.851} \\
    
    & Mixtral     
    & 0.938 & \mgtScore{0.845} & \mgtScore{0.837} & \mgtScore{0.851} & \mgtScore{0.845} & \mgtScore{0.801} & \mgtScore{0.771} & \mgtBest{0.880}
    & 0.938 & \mgtScore{0.841} & \mgtScore{0.830} & \mgtScore{0.833} & \mgtScore{0.758} & \mgtScore{0.785} & \mgtScore{0.773} & \mgtBest{0.878} \\
    
    & Moonshot    
    & 0.931 & \mgtScore{0.847} & \mgtScore{0.843} & \mgtScore{0.840} & \mgtScore{0.816} & \mgtScore{0.780} & \mgtScore{0.792} & \mgtBest{0.868}
    & 0.922 & \mgtScore{0.835} & \mgtScore{0.829} & \mgtScore{0.827} & \mgtScore{0.788} & \mgtScore{0.784} & \mgtScore{0.778} & \mgtBest{0.862} \\
    
    & Llama-3.1     
    & 0.920 & \mgtScore{0.820} & \mgtScore{0.806} & \mgtScore{0.814} & \mgtScore{0.811} & \mgtScore{0.761} & \mgtScore{0.751} & \mgtBest{0.854}
    & 0.912 & \mgtScore{0.785} & \mgtScore{0.790} & \mgtScore{0.784} & \mgtScore{0.753} & \mgtScore{0.750} & \mgtScore{0.739} & \mgtBest{0.853} \\
    
    & GPT-4o-mini 
    & 0.905 & \mgtScore{0.830} & \mgtScore{0.810} & \mgtScore{0.838} & \mgtScore{0.810} & \mgtScore{0.768} & \mgtScore{0.790} & \mgtBest{0.868}
    & 0.898 & \mgtScore{0.834} & \mgtScore{0.823} & \mgtScore{0.823} & \mgtScore{0.829} & \mgtScore{0.795} & \mgtScore{0.765} & \mgtBest{0.867} \\
    \cmidrule(lr){2-18}
    & Average     
    & 0.920 & \mgtScore{0.835} & \mgtScore{0.828} & \mgtScore{0.836} & \mgtScore{0.825} & \mgtScore{0.777} & \mgtScore{0.775} & \mgtBest{0.867}
    & 0.912 & \mgtScore{0.823} & \mgtScore{0.817} & \mgtScore{0.814} & \mgtScore{0.776} & \mgtScore{0.779} & \mgtScore{0.764} & \mgtBest{0.862} \\
    \midrule
    
    \multirow{6}{*}{STEM}
    & GPT-3.5     
    & 0.946 & \mgtScore{0.882} & \mgtScore{0.880} & \mgtScore{0.883} & \mgtScore{0.867} & \mgtScore{0.802} & \mgtScore{0.805} & \mgtBest{0.914}
    & 0.942 & \mgtScore{0.847} & \mgtScore{0.850} & \mgtScore{0.855} & \mgtScore{0.847} & \mgtScore{0.808} & \mgtScore{0.801} & \mgtBest{0.907} \\
    
    & Mixtral     
    & 0.964 & \mgtScore{0.887} & \mgtScore{0.883} & \mgtScore{0.878} & \mgtScore{0.874} & \mgtScore{0.830} & \mgtScore{0.809} & \mgtBest{0.912}
    & 0.966 & \mgtScore{0.879} & \mgtScore{0.877} & \mgtScore{0.875} & \mgtScore{0.867} & \mgtScore{0.813} & \mgtScore{0.822} & \mgtBest{0.919} \\
    
    & Moonshot    
    & 0.959 & \mgtScore{0.895} & \mgtScore{0.880} & \mgtScore{0.891} & \mgtScore{0.885} & \mgtScore{0.841} & \mgtScore{0.830} & \mgtBest{0.916}
    & 0.958 & \mgtScore{0.893} & \mgtScore{0.885} & \mgtScore{0.888} & \mgtScore{0.892} & \mgtScore{0.834} & \mgtScore{0.853} & \mgtBest{0.917} \\
    
    & Llama-3.1     
    & 0.959 & \mgtScore{0.852} & \mgtScore{0.844} & \mgtScore{0.851} & \mgtScore{0.757} & \mgtScore{0.809} & \mgtScore{0.801} & \mgtBest{0.902}
    & 0.956 & \mgtScore{0.860} & \mgtScore{0.852} & \mgtScore{0.852} & \mgtScore{0.856} & \mgtScore{0.809} & \mgtScore{0.822} & \mgtBest{0.910} \\
    
    & GPT-4o-mini 
    & 0.942 & \mgtScore{0.888} & \mgtScore{0.882} & \mgtScore{0.885} & \mgtScore{0.780} & \mgtScore{0.826} & \mgtScore{0.827} & \mgtBest{0.913}
    & 0.945 & \mgtScore{0.854} & \mgtScore{0.862} & \mgtScore{0.890} & \mgtScore{0.888} & \mgtScore{0.837} & \mgtScore{0.831} & \mgtBest{0.919} \\
    \cmidrule(lr){2-18}
    & Average     
    & 0.954 & \mgtScore{0.881} & \mgtScore{0.874} & \mgtScore{0.878} & \mgtScore{0.832} & \mgtScore{0.822} & \mgtScore{0.814} & \mgtBest{0.911}
    & 0.953 & \mgtScore{0.867} & \mgtScore{0.865} & \mgtScore{0.872} & \mgtScore{0.870} & \mgtScore{0.820} & \mgtScore{0.826} & \mgtBest{0.914} \\
    \midrule
    
    \multirow{6}{*}{Humanities}
    & GPT-3.5     
    & 0.903 & \mgtScore{0.854} & \mgtScore{0.856} & \mgtScore{0.831} & \mgtScore{0.863} & \mgtScore{0.816} & \mgtScore{0.780} & \mgtBest{0.881}
    & 0.898 & \mgtScore{0.827} & \mgtScore{0.824} & \mgtScore{0.834} & \mgtScore{0.826} & \mgtScore{0.787} & \mgtScore{0.768} & \mgtBest{0.873} \\
    
    & Mixtral     
    & 0.954 & \mgtScore{0.856} & \mgtScore{0.854} & \mgtScore{0.851} & \mgtScore{0.850} & \mgtScore{0.817} & \mgtScore{0.805} & \mgtBest{0.876}
    & 0.952 & \mgtScore{0.831} & \mgtScore{0.828} & \mgtScore{0.838} & \mgtScore{0.832} & \mgtScore{0.803} & \mgtScore{0.791} & \mgtBest{0.870} \\
    
    & Moonshot    
    & 0.936 & \mgtScore{0.851} & \mgtScore{0.840} & \mgtScore{0.843} & \mgtScore{0.844} & \mgtScore{0.779} & \mgtScore{0.807} & \mgtBest{0.862}
    & 0.945 & \mgtScore{0.825} & \mgtScore{0.829} & \mgtScore{0.841} & \mgtScore{0.823} & \mgtScore{0.789} & \mgtScore{0.814} & \mgtBest{0.872} \\
    
    & Llama-3.1     
    & 0.928 & \mgtScore{0.809} & \mgtScore{0.800} & \mgtScore{0.806} & \mgtScore{0.810} & \mgtScore{0.771} & \mgtScore{0.771} & \mgtBest{0.851}
    & 0.922 & \mgtScore{0.701} & \mgtScore{0.787} & \mgtScore{0.795} & \mgtScore{0.789} & \mgtScore{0.739} & \mgtScore{0.768} & \mgtBest{0.856} \\
    
    & GPT-4o-mini 
    & 0.903 & \mgtScore{0.843} & \mgtScore{0.836} & \mgtScore{0.848} & \mgtScore{0.838} & \mgtScore{0.799} & \mgtScore{0.771} & \mgtBest{0.868}
    & 0.905 & \mgtScore{0.830} & \mgtScore{0.747} & \mgtScore{0.830} & \mgtScore{0.806} & \mgtScore{0.806} & \mgtScore{0.730} & \mgtBest{0.868} \\
    \cmidrule(lr){2-18}
    & Average     
    & 0.925 & \mgtScore{0.843} & \mgtScore{0.837} & \mgtScore{0.836} & \mgtScore{0.841} & \mgtScore{0.796} & \mgtScore{0.787} & \mgtBest{0.867}
    & 0.924 & \mgtScore{0.803} & \mgtScore{0.803} & \mgtScore{0.828} & \mgtScore{0.815} & \mgtScore{0.785} & \mgtScore{0.774} & \mgtBest{0.868} \\
    \midrule
    
    \multirow{6}{*}{AIGTBench}
    & GPT-3.5     
    & 0.964 & \mgtScore{0.875} & \mgtScore{0.870} & \mgtScore{0.876} & \mgtScore{0.863} & \mgtScore{0.840} & \mgtScore{0.804} & \mgtBest{0.894}
    & 0.959 & \mgtScore{0.873} & \mgtScore{0.847} & \mgtScore{0.863} & \mgtScore{0.851} & \mgtScore{0.833} & \mgtScore{0.814} & \mgtBest{0.892} \\
    
    & GPT-4o-mini 
    & 0.956 & \mgtScore{0.823} & \mgtScore{0.833} & \mgtScore{0.866} & \mgtScore{0.800} & \mgtScore{0.855} & \mgtScore{0.820} & \mgtBest{0.896}
    & 0.954 & \mgtScore{0.798} & \mgtScore{0.835} & \mgtScore{0.834} & \mgtScore{0.802} & \mgtScore{0.815} & \mgtScore{0.803} & \mgtBest{0.889} \\
    
    & Llama-1     
    & 0.913 & \mgtScore{0.892} & \mgtScore{0.867} & \mgtScore{0.897} & \mgtScore{0.896} & \mgtScore{0.851} & \mgtScore{0.878} & \mgtBest{0.915}
    & 0.916 & \mgtScore{0.881} & \mgtScore{0.870} & \mgtScore{0.873} & \mgtScore{0.873} & \mgtScore{0.845} & \mgtScore{0.794} & \mgtBest{0.910} \\
    
    & Llama-2     
    & 0.918 & \mgtScore{0.856} & \mgtScore{0.817} & \mgtScore{0.855} & \mgtScore{0.837} & \mgtScore{0.795} & \mgtScore{0.775} & \mgtBest{0.892}
    & 0.919 & \mgtScore{0.844} & \mgtScore{0.840} & \mgtScore{0.856} & \mgtScore{0.844} & \mgtScore{0.794} & \mgtScore{0.811} & \mgtBest{0.893} \\
    
    & Llama-3.1     
    & 0.915 & \mgtScore{0.864} & \mgtScore{0.842} & \mgtScore{0.856} & \mgtScore{0.837} & \mgtScore{0.836} & \mgtScore{0.777} & \mgtBest{0.899}
    & 0.913 & \mgtScore{0.841} & \mgtScore{0.762} & \mgtScore{0.840} & \mgtScore{0.837} & \mgtScore{0.796} & \mgtScore{0.796} & \mgtBest{0.895} \\
    \cmidrule(lr){2-18}
    & Average     
    & 0.933 & \mgtScore{0.862} & \mgtScore{0.846} & \mgtScore{0.870} & \mgtScore{0.847} & \mgtScore{0.835} & \mgtScore{0.811} & \mgtBest{0.899}
    & 0.932 & \mgtScore{0.848} & \mgtScore{0.831} & \mgtScore{0.853} & \mgtScore{0.842} & \mgtScore{0.817} & \mgtScore{0.804} & \mgtBest{0.896} \\
    \midrule
    
    \multicolumn{2}{l}{Overall Average}
    & 0.933 & \mgtScore{0.855} & \mgtScore{0.846} & \mgtScore{0.855} & \mgtScore{0.836} & \mgtScore{0.807} & \mgtScore{0.797} & \mgtBest{0.886}
    & 0.930 & \mgtScore{0.835} & \mgtScore{0.829} & \mgtScore{0.842} & \mgtScore{0.826} & \mgtScore{0.800} & \mgtScore{0.792} & \mgtBest{0.885} \\
    \bottomrule
    \end{tabular}
    }
\end{table*}

\mypara{Baselines \& Implementation}
We follow the experimental protocol of~\citet{DBLP:conf/kdd/LiuZL0ZWGT00025} and compare our method against several representative baselines.
We include the classical methods LwF~\cite{li2017learning}, iCaRL~\cite{rebuffi2017icarl}, and BiC~\cite{wu2019large}, as well as the more recent EASE~\cite{zhou2024expandable}, PASS~\cite{zhu2021prototype} and SimpleCIL~\cite{zhou2025revisiting}.
Following~\citet{DBLP:conf/kdd/LiuZL0ZWGT00025}, we equip LwF, iCaRL, BiC, and EASE with a replay buffer of 100 stored exemplars per old class, whereas PASS and SimpleCIL remain exemplar-free.
As for the backbones, we adopt DeBERTa-v3-base~\cite{he2021debertav3} and RoBERTa-base~\cite{liu2019roberta} as the text encoders for feature extraction.

\mypara{Evaluation Metrics}
All methods use the same data splits and incremental protocols. \method stores no raw exemplars, and its statistical-state memory is analyzed separately in \Cref{tab:compression_ablation}.
As the primary evaluation metric, we use macro-F1.
To provide a more detailed analysis of model behavior, we report not only the overall macro-F1 over all seen classes, but also separate scores on old classes and new classes at each incremental stage.
This multi-dimensional evaluation enables us to quantify how different methods balance two competing objectives: mitigating catastrophic forgetting on old classes and adapting to newly introduced ones.
Unless otherwise specified, we report the mean performance over three random seeds.

\begin{figure*}[t]
    \centering
    \includegraphics[width=0.94\textwidth]{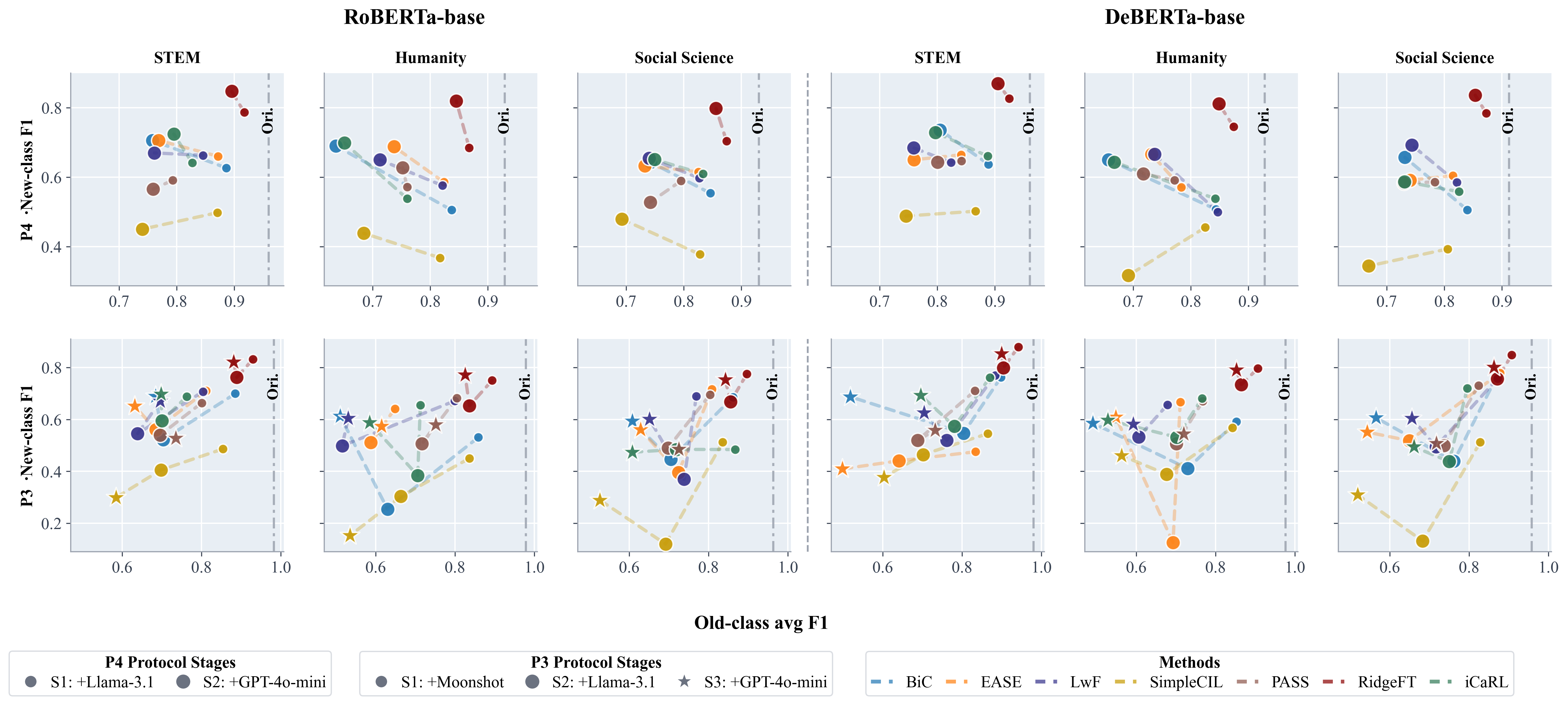}
    \caption{
    Experiments on academic topics.
    P3 starts with $3$ initial classes and sequentially adds Moonshot, Llama-3.1, and GPT-4o-mini;
    P4 starts with $4$ initial classes and sequentially adds Llama-3.1 and GPT-4o-mini.
    }
    \label{fig:trajectory_academic}
\end{figure*}

\section{Experiments}
\subsection{Comparison with Baselines}
We comprehensively compare \method with a wide range of baselines under $3$ evaluation protocols.
Baselines marked with $\ast$, such as LwF$\ast$, iCaRL$\ast$, BiC$\ast$, and EASE$\ast$, denote methods equipped with a replay mechanism that stores historical samples.

As shown in \Cref{fig:trajectory_academic,fig:trajectory_social_media}, \method consistently outperforms all baselines across the P3 and P4 incremental stages in both domains.
In the academic setting, encompassing the STEM, Humanities, and Social Science datasets, under P3 (sequentially adding Moonshot, Llama-3.1, and GPT-4o-mini), \method maintains strong old-class F1 (0.913, 0.870, 0.861) and new-class F1 (0.814, 0.729, 0.799) across the three stages.
Under P4, it achieves full-F1 scores of 0.862 and 0.838, significantly outpacing the strongest baseline, which drops from 0.797 to 0.730.
This highlights that replay-based baselines still struggle to balance old and new classes, while replay-free baselines such as PASS and SimpleCIL are even more limited when generator classes arrive sequentially.
This robust advantage extends to the social media setting, where \method sustains high full-F1 scores across the P3 stages (0.945, 0.868, 0.855) and P4 stages (0.878, 0.867). 
Notably, at the final social media P4 stage, \method's new-class F1 reaches 0.899, substantially beating the strongest baseline (0.735).
These results confirm that \method remains highly stable when absorbing successive generator arrivals without compromising past knowledge.

This advantage is further confirmed under the P5 protocol (refer to~\Cref{tab:P5_full_f1}).
Averaged over the two backbones, \method achieves 0.886 full-F1, 0.902 old-class F1, and 0.804 new-class F1.
Compared with the strongest full-F1 baseline, it improves full-F1 by 0.037.
Compared with the best baseline on new-class F1, it improves new-class F1 by 0.107, while also outperforming the best old-class baseline on old-class F1.
This clearly reveals a core limitation of baselines: they are better at preserving the decision boundaries of previously learned classes, yet remain less effective at efficiently absorbing information from new classes.
In contrast, \method not only performs better in retaining old classes, but also delivers a substantial improvement on the more critical task of recognizing new classes, surpassing the strongest baseline by 0.107. Complete per-domain and per-backbone new- and old-class F1 are reported in \Cref{tab:newclass_full,tab:oldclass_full}.

\begin{figure}[h!]
    \centering
    \includegraphics[width=0.70\linewidth]{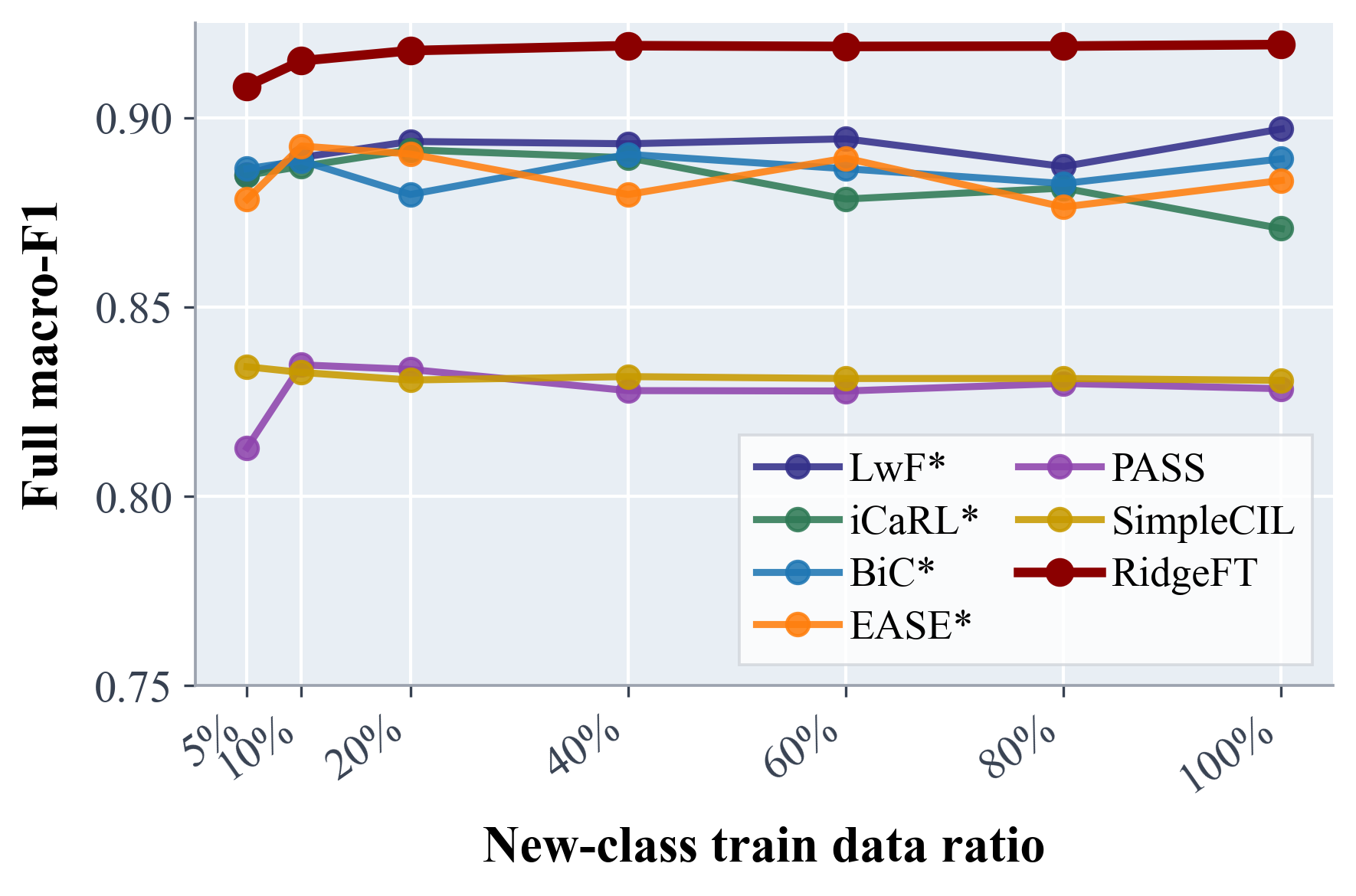}
    \caption{Full-F1 under varying target-class data proportions.}
    \label{fig:ratio}
\end{figure}

Finally, using GPT-4o-mini as the target generator (STEM / DeBERTa-base / P5), we analyze the impact of target-class data proportions. As shown in \Cref{fig:ratio}, when reducing target-class data from 100\% to 5\%, \method's full-F1 remains highly stable (ranging from 0.908 to 0.919).
Conversely, the strongest replay baseline (LwF$\ast$) averages only 0.893, and the replay-free SimpleCIL drops to 0.831. 
This confirms that \method does not rely on abundant new-class data, demonstrating superior data efficiency and robustness even under severe low-resource incremental conditions.

To further assess scalability and generalization, we evaluate \method on a chronological cross-domain stream and on class sets of up to 18 source classes (\Cref{app:scaling}), and on the independently constructed RAID benchmark~\cite{DBLP:conf/acl/DuganHTZLXIC24}, under DIPPER paraphrasing attacks, and through a misattribution analysis (\Cref{app:extra}). \method degrades gradually rather than abruptly as the stream and class set grow, and its advantage over baselines holds on the independent benchmark.

\begin{figure*}[t]
    \centering
    \includegraphics[width=0.95\linewidth]{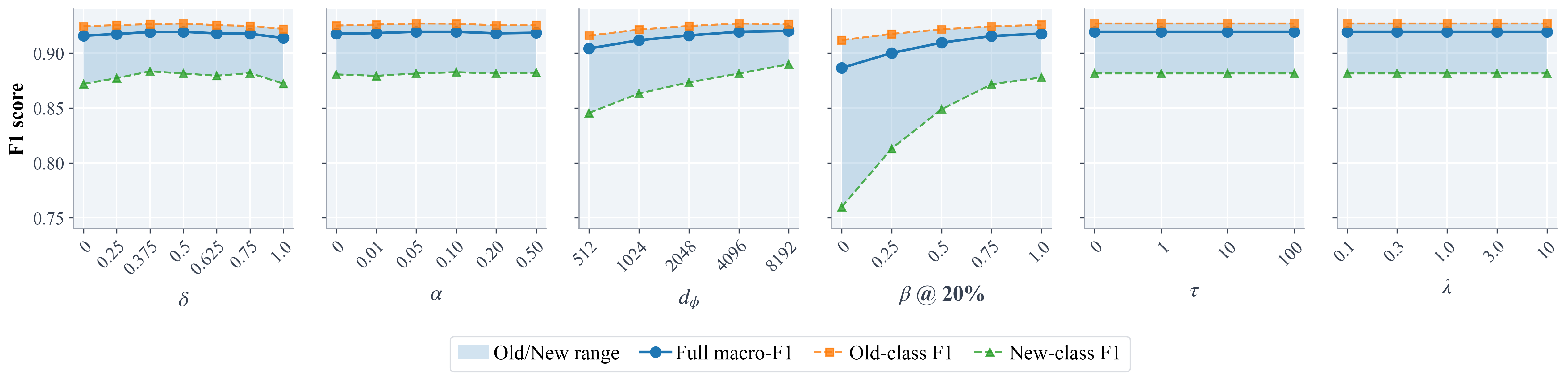}
    \caption{Parameter sensitivity of \method. We vary one hyperparameter at a time while keeping the others fixed, including the covariance calibration exponent $\delta$, trace shrinkage coefficient $\alpha$, random feature dimension $d_\phi$, class-reweighting strength $\beta$ under the 20\% setting, smoothing constant $\tau$, and ridge regularization coefficient $\lambda$.}
    \label{fig:ablation}
\end{figure*}

\subsection{Ablation Studies}

\begin{table}[h!]
\centering
\caption{Component ablation of \method under different new-class data ratios. 
W denotes whitening, RFL denotes random feature lift, and CBR denotes class-balanced ridge. 
``Full'' denotes full macro-F1 over all seen classes, while ``New'' denotes new-class F1.}
\label{tab:ridgeft-component-ablation}
\resizebox{0.48\textwidth}{!}{
\begin{tabular}{lcccccccc}
\toprule
\multirow{2}{*}{Ratio}
& \multicolumn{2}{c}{Only W}
& \multicolumn{2}{c}{Only RFL}
& \multicolumn{2}{c}{Only CBR}
& \multicolumn{2}{c}{\method} \\
\cmidrule(lr){2-3}
\cmidrule(lr){4-5}
\cmidrule(lr){6-7}
\cmidrule(lr){8-9}
& Full & New & Full & New & Full & New & Full & New \\
\midrule
20\%  & 0.881 & 0.741 & 0.887 & 0.763 & 0.912 & 0.865 & \textbf{0.918} & \textbf{0.878} \\
40\%  & 0.904 & 0.830 & 0.908 & 0.839 & 0.912 & 0.865 & \textbf{0.919} & \textbf{0.880} \\
60\%  & 0.909 & 0.851 & 0.913 & 0.861 & 0.913 & 0.865 & \textbf{0.919} & \textbf{0.881} \\
80\%  & 0.912 & 0.861 & 0.916 & 0.872 & 0.912 & 0.865 & \textbf{0.919} & \textbf{0.880} \\
100\% & 0.910 & 0.858 & 0.917 & 0.878 & 0.912 & 0.863 & \textbf{0.919} & \textbf{0.881} \\
\bottomrule
\end{tabular}
}
\end{table}

\mypara{Component-wise Ablation}
We ablate the $3$ core components of \method using GPT-4o-mini as the target generator (STEM / DeBERTa-base / P5).
As shown in \Cref{tab:ridgeft-component-ablation}, each component contributes distinctly, and they are highly complementary.
When Whitening is used alone, the model can stabilize encoder features to some extent, but its adaptability remains limited under low-resource conditions.
For example, with only $20\%$ of the new-class data, the new-class F1 reaches only $0.741$.
In contrast, Random Feature Lift consistently outperforms Whitening across all data ratios, indicating that random nonlinear features enhance the separability of different generators.
Class-Balanced Ridge (CBR) is the strongest individual component.
Even with only $20\%$ of the data, it achieves a new-class F1 of $0.865$, suggesting that it plays a central role in mitigating the bias between old and new classes.
Notably, both CBR and the full \method exhibit stability across data ratios ranging from $20\%$ to $100\%$.
According to \Cref{eq:class_balance_weight}, when $\beta=1$ and $\tau=0$, the class weight satisfies $\omega_c \propto 1/N_c$, which exactly offsets the linear growth of class statistics ($A_c, q_c$) with respect to the sample size $N_c$.
Consequently, the ridge update effectively operates on class-averaged statistics, rendering the model largely insensitive to the sampling ratio once stable estimation is reached.
Finally, combining all three components yields the best overall performance.
Their complementarity is particularly evident in the $20\%$ low-resource setting, where the full \method boosts the new-class F1 from $0.865$ (CBR alone) to 0.878.
Additional frozen-representation analysis is provided in~\Cref{app:repr_sufficiency}.

\mypara{Parameter Sensitivity Ablation}
\Cref{fig:ablation} presents the hyperparameter sensitivity of \method.
Overall, it exhibits strong robustness to most hyperparameters.
For example, varying the smoothing constant $\tau$ and the ridge regularization coefficient $\lambda$ causes almost no performance fluctuation, with full-F1 remaining consistently around $0.919$.
From the perspective of feature stabilization, a moderate amount of trace shrinkage, with $\alpha \in [0.05, 0.10]$, is already sufficient to ensure reliable covariance estimation.
The covariance calibration exponent $\delta$ performs best in the range $[0.375, 0.625]$, with full-F1 reaching $0.919$ at $\delta=0.5$.
However, $\delta=1.0$, which corresponds to inverse-covariance scaling stronger than standard whitening, instead suppresses useful discriminative information and leads to a performance drop.
In terms of performance gains, the random feature dimension $d_\phi$ and the class reweighting strength $\beta$ play the most important roles.
As $d_\phi$ increases from $512$ to $8192$, the higher-dimensional feature space substantially improves the separability of new generators, steadily raising full-F1 and new-class F1 to $0.920$ and $0.890$, respectively.
In addition, under the low-resource setting with $20\%$ data, strong reweighting with $\beta=1.0$ boosts new-class F1 by nearly 12 percentage points compared with the unweighted case $\beta=0$, increasing it from $0.760$ to $0.878$.
This highlights its decisive role in alleviating the particularly severe imbalance between old and new classes.

\begin{table}[h!]
\centering
\caption{
Sufficient-statistic compression ablation (default setting: $d_\phi = 4096$, 6 classes). Per-class schemes retain update flexibility, while merged schemes minimize storage by assuming future updates only add new classes, not new samples to old ones.}
\label{tab:compression_ablation}
\small
\resizebox{0.48\textwidth}{!}{\begin{tabular}{lcccc}
\toprule
\textbf{Storage scheme} & \textbf{Stored $A$ statistics} & \textbf{Storage} & \textbf{Compression} & \textbf{full-F1} \\
\midrule
fp64 baseline 
& $6 \times A_c$ in fp64 
& 782.35 MiB 
& $1.00\times$ 
& 0.919 \\
fp32 per-class 
& $6 \times A_c$ in fp32 
& 398.35 MiB 
& $1.96\times$ 
& 0.919 \\
bf16 per-class 
& $6 \times A_c$ in bf16 
& 206.35 MiB 
& $3.79\times$ 
& 0.920 \\
merged fp32 
& $1 \times \bar A$ in fp32 
& 78.35 MiB 
& $9.99\times$ 
& 0.919 \\
\textbf{merged bf16} 
& $\mathbf{1 \times \bar A}$ \textbf{in bf16} 
& \textbf{46.35 MiB} 
& $\mathbf{16.9\times}$ 
& \textbf{0.920} \\
\bottomrule
\end{tabular}}

\end{table}

\mypara{Sufficient-Statistic Compression Ablation}
Although \method is exemplar-free, it is not memory-free.
Its closed-form update still requires storing the statistics needed for ridge regression: the calibration parameters, the fixed projection matrix $R$, and the class-wise sufficient statistics $A_c$, $q_c$, and $N_c$.
The high-dimensional $A_c \in \mathbb{R}^{d_\phi \times d_\phi}$ dominates this storage cost.
Under the default setting ($d_\phi=4096$, $6$ classes), the fp64 implementation requires $782.35$ MiB in total, with the six class-specific $A_c$ matrices alone accounting for $768$ MiB.

To reduce this bottleneck, we examine compression along two orthogonal dimensions. The first is numerical precision (fp32 or bf16), which preserves the full class-wise statistic table.
This retains the flexibility to reweight classes post-training or incorporate new samples into previously seen classes. The second is class aggregation, which stores only the weighted merged matrix $\bar A=\sum_c \omega_c A_c$.
This significantly reduces the memory footprint but folds the current class weights into the matrix, sacrificing the ability to update old classes individually.
Importantly, this limitation is fully compatible with our P3/P4/P5 protocols, where incremental stages only introduce new generator classes.

As shown in \Cref{tab:compression_ablation}, \method is highly robust to low-precision storage.
Reducing per-class $A_c$ to fp32 or bf16 substantially cuts storage while keeping the full-F1 consistently at or above 0.919.
Class aggregation yields even larger gains by removing redundant class-specific matrices.
Among all variants, merged bf16 provides the optimal trade-off: it requires only 46.35 MiB (a $16.9\times$ compression ratio) while achieving a full-F1 of 0.920.
Overall, \method should be viewed as replay-free rather than storage-free.
Merged bf16 is the most compact choice for our new-class-only protocols, whereas per-class bf16 is preferable if future updates may add samples to old classes.

\section{Conclusion}
This paper addresses the challenge of lifelong MGT attribution: continuously learning new generators without suffering catastrophic forgetting or relying on historical text replay.
To overcome the dilemma between representation drift (from fine-tuning) and memory overhead (from exemplars), we propose \method.
By freezing a task-tuned encoder and formulating incremental adaptation as closed-form ridge regression, our replay-free framework balances old and new classes without encoder parameter updates.
Extensive experiments confirm that \method achieves a superior trade-off against baselines, demonstrating that combining stable representations with analytic updates offers a scalable path for future MGT attribution systems.

\section*{Limitations}
Although \method achieves strong performance in MGT attribution, it still has several limitations.

\begin{itemize}
    \item \textbf{Storage Requirements and Compression:}
    \method is replay-free but not storage-free: its closed-form updates still store the calibration parameters, the fixed projection matrix, and the class-wise sufficient statistics. Our compression study cuts this cost substantially, and further compressing these statistics is left to future work.

    \item \textbf{Long-term Deployment and Representation Stability:}
    Because \method keeps the encoder frozen during the incremental stage, it cannot absorb an unlimited number of generators without eventual degradation: as future generators diverge from the initial ones, the frozen representation may lack the cues that later updates need. Periodic encoder refresh between long stretches of low-cost analytic updates is therefore a promising direction (see \Cref{app:scaling}).

    \item \textbf{Adversarial Robustness:}
    Because \method attributes text in a frozen representation space, paraphrasing attacks that rewrite surface style can degrade its attribution accuracy (see \Cref{app:extra}). Improving the robustness of the base-stage encoder before freezing remains an important direction for future work.
\end{itemize}

\section*{Ethical Considerations}
This work focuses on lifelong machine-generated text attribution under benchmark settings.
It does not involve private user data, human subjects, or new generative or attack capabilities. The main deployment risk is misattribution, which may lead to incorrect judgments about text sources.
We therefore recommend using MGT attribution models as auxiliary evidence rather than the sole basis for high-stakes decisions, while also considering domain shift, transparency, and human oversight.

\section*{Acknowledgments}
We sincerely thank the anonymous reviewers, the Area Chairs (AC), the Senior Area Chairs (SAC), and the Program Chairs for their careful reading, constructive comments, and thoughtful handling of our submission throughout the ACL Rolling Review and EMNLP 2026 process, which greatly helped us improve this paper.
This work was supported in part by the National Key Research and Development Program of China (No.~2025YFB3110200), the Guangdong Basic and Applied Basic Research Foundation (No.~2026A1515030046), the State Key Laboratory of Internet Architecture, Tsinghua University (No.~HLW2025ZD14), the Yangcheng Scholars Research Project (No.~2024312049), and the Guangdong Provincial Key Lab of Integrated Communication, Sensing and Computation for Ubiquitous Internet of Things (No.~2023B1212010007).

\bibliographystyle{plain}
\bibliography{custom}

\appendix
\setcounter{figure}{0}
\renewcommand{\thefigure}{A\arabic{figure}}
\setcounter{table}{0}
\renewcommand{\thetable}{A\arabic{table}}

\section{Protocol and Metric Details}
\label{app:protocol}
The attribution pool contains six classes, one Human source and five generators. Protocol P$k$ starts from $k$ initial classes and introduces the remaining $6-k$ generators one per incremental step, so P5 adds a single held-out generator (rotated across generators in \Cref{tab:P5_full_f1}), while P4 and P3 add two and three generators in sequence. The academic arrival orders are given in \Cref{fig:trajectory_academic} and the social-media orders in \Cref{fig:trajectory_social_media}. At every step, predictions are made over all classes seen so far, and we report three quantities: full-F1, the macro-F1 over all seen classes; old-class F1, the macro-F1 over the initial $k$ classes; and new-class F1, the F1 of the single generator introduced at the current step.

\section{Social Media Incremental Trajectories}
\label{app:social_trajectories}
\begin{figure}[h!]
    \centering
    \includegraphics[width=0.5\textwidth]{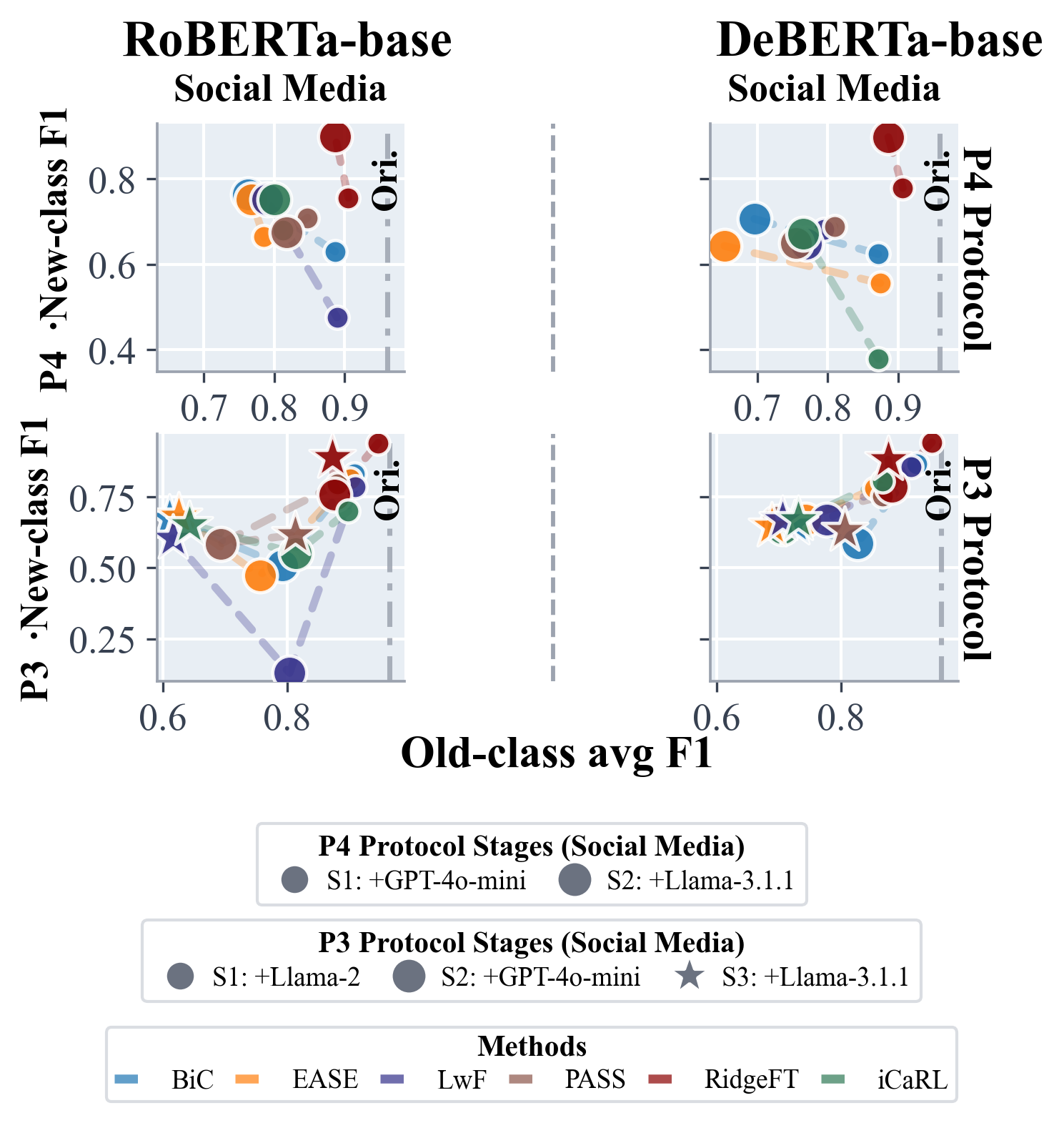}
    \caption{
Lifelong MGT attribution experiments on the social media topic.
P3 starts with three initial classes and sequentially adds Llama-2, GPT-4o-mini, and Llama-3.1;
P4 starts with four initial classes and sequentially adds GPT-4o-mini and Llama-3.1.
}
    \label{fig:trajectory_social_media}
\end{figure}

\Cref{fig:trajectory_social_media} reports the full P3 and P4 attribution trajectories on the social media domain, complementing the academic trajectories shown in \Cref{fig:trajectory_academic} of the main text.

\section{Sufficiency Analysis of Frozen Representations}
\label{app:repr_sufficiency}

Since \method freezes the task-tuned encoder during the incremental stage, a natural question arises: do frozen representations still preserve discriminative information for future generators?
To answer this question, we compare three feature spaces derived from the same frozen encoder: the raw representation $h=f_\theta(x)$, the covariance-calibrated representation $\tilde h=P_\delta(h-\mu)$, and the random feature representation $z=\mathrm{LN}(\mathrm{ReLU}(R\tilde h))$.
We diagnose these feature spaces using nearest-centroid probes and ridge probes, where none of the methods updates the encoder parameters.

\begin{figure*}[h!]
    \centering
    \includegraphics[width=\linewidth]{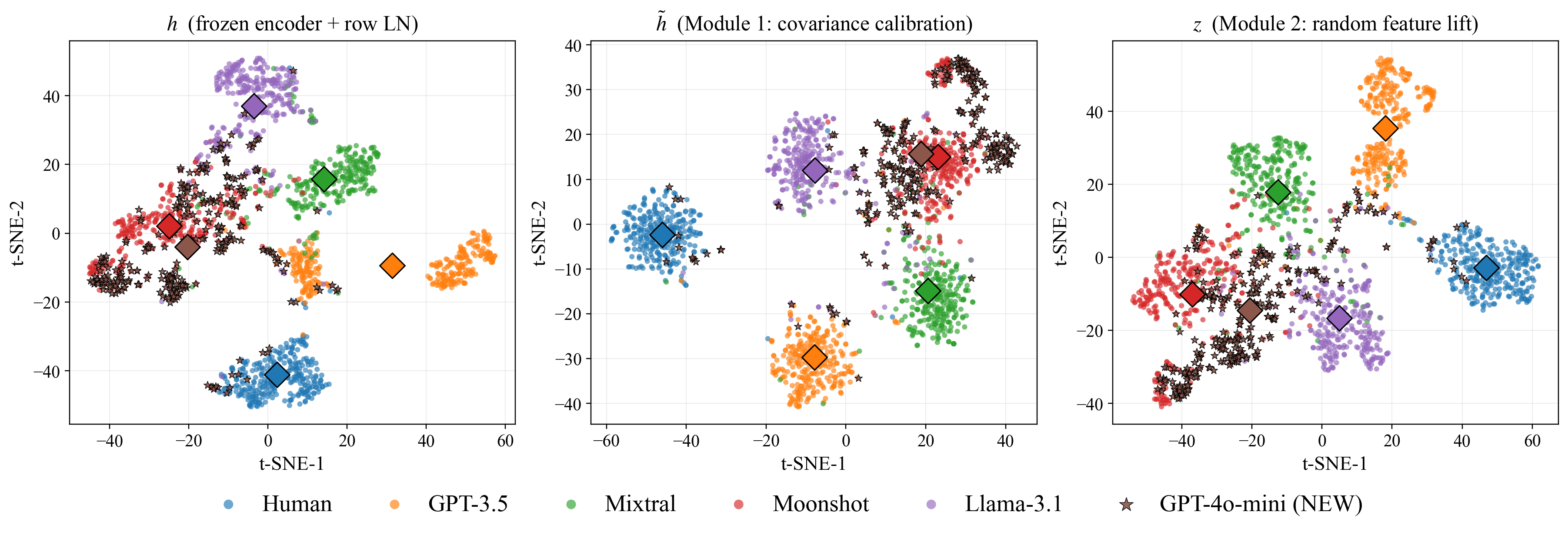}
    \caption{
    t-SNE visualization of different frozen feature spaces. We compare the raw frozen representation $h$, the calibrated representation $\tilde h$, and the random feature representation $z$. The figure shows that the newly introduced generator does not completely collapse into the old classes, and that the feature transformations in \method make the class structure more clearly separated (STEM / DeBERTa-base / P5 setting).
    }
    \label{fig:tsne_feature_spaces}
\end{figure*}

\begin{table}[h!]
\centering
\small
\caption{
Sufficiency analysis of frozen representations. All methods use the same frozen encoder. Ridge on $h$ already achieves a relatively strong new-class F1, suggesting that the frozen representation retains discriminative information for future generators. \method further improves performance, indicating that covariance calibration and random feature mapping can exploit this information more effectively.
}
\label{tab:repr_sufficiency}
\resizebox{0.48\textwidth}{!}{
\begin{tabular}{lccc}
\toprule
Probe on frozen-encoder features & Full-F1 & Old-F1 & New-F1 \\
\midrule
Nearest-centroid on $h$ & 0.788 & 0.847 & 0.495 \\
Nearest-centroid on $z$ & 0.854 & 0.879 & 0.730 \\
Ridge on $h$ & 0.887 & 0.895 & 0.844 \\
\method on $z$ & \textbf{0.919} & \textbf{0.927} & \textbf{0.881} \\
\bottomrule
\end{tabular}}
\end{table}

This analysis is conducted under the P5 protocol, where the model is first trained on five initial classes and GPT-4o-mini is introduced as the final new generator.
As shown in \Cref{tab:repr_sufficiency}, applying a ridge probe directly on the raw frozen representation $h$ already yields a new-class F1 of $0.844$, which suggests that the frozen encoder is not limited to the initial classes and still preserves useful discriminative cues for subsequent generators.
Building on this, \method further improves the new-class F1 to $0.881$, while also maintaining stable old-class F1.
On the other hand, the nearest-centroid probe improves from $0.495$ on $h$ to $0.730$ on $z$, indicating that the feature transformations in \method make the class structure more suitable for statistic-based analytic classification, consistent with the t-SNE visualization in \Cref{fig:tsne_feature_spaces}.

It should be noted that this analysis does not imply that a frozen encoder is always sufficient for arbitrary future generators.
The random feature mapping can only enhance the usability of the information already present in the frozen representation, but it cannot recover generator-specific information that is entirely absent from the encoder.
Therefore, \method is better suited to scenarios where the frozen representation still provides at least partial coverage of future generators. For generators that are extremely out of distribution, periodic adaptation of the encoder may still be necessary.

\section{Longer Streams and Larger Class Counts}
\label{app:scaling}

\mypara{Chronological cross-domain stream}
To examine how \method behaves over a longer and temporally ordered stream, we merge MGT-Academic and AIGTBench into a single chronological benchmark. The stream has 8 classes (Human and 7 generators): every model shared by the two sources is merged into one class with exact within-class deduplication, and the classes are ordered by release date. We balance each class to 6{,}000 texts with an 80/10/10 train/validation/test split. The four cross-source classes break the spurious source-to-label correspondence, while the single-source classes keep their original domain composition, so the stream deliberately couples cross-generational with cross-domain shift. The encoder is fine-tuned only on the four earliest sources, Human, Llama-1 (Feb.\ 2023), GPT-3.5 Turbo (Mar.\ 2023), and Llama-2 (Jul.\ 2023), and is then frozen. Mixtral-8x7B (Dec.\ 2023), Moonshot-v1 (Feb.\ 2024), GPT-4o-mini (Jul.\ 2024), and Llama-3.1 (Jul.\ 2024) are added in chronological order, spanning about 17 months. Old-class F1 is the macro-F1 over the four initial classes, new-class F1 is the F1 of the latest arriving class, and full-F1 is the macro-F1 over all seen classes.
As shown in \Cref{tab:chrono_stream}, full-F1 decreases smoothly from 0.956 to 0.795 with DeBERTa-v3-base while the initial-class macro-F1 decreases more slowly, ending at 0.888, so most of the drop concentrates in the accumulated incremental classes. Replacing the backbone with DeBERTa-v3-large raises the final full-F1, old-class F1, and new-class F1 from 0.795/0.888/0.702 to 0.812/0.895/0.729, indicating that a stronger initial encoder improves the attainable performance for later generators while preserving the same gradual trend.

\begin{table}[h!]
\centering

\caption{Chronological cross-domain stream combining MGT-Academic and AIGTBench. The encoder is frozen after the four initial classes, and four generators are added in temporal order. Full/Old/New denote full macro-F1, initial-class macro-F1, and latest-class F1.}
\label{tab:chrono_stream}
\resizebox{0.48\textwidth}{!}{
\begin{tabular}{lccccccc}
\toprule
\multirow{2}{*}{Incremental step} & \multirow{2}{*}{Seen} & \multicolumn{3}{c}{DeBERTa-v3-base} & \multicolumn{3}{c}{DeBERTa-v3-large} \\
\cmidrule(lr){3-5}\cmidrule(lr){6-8}
 & & Full & Old & New & Full & Old & New \\
\midrule
Initial       & 4 & 0.956 & 0.956 & --    & 0.960 & 0.960 & --    \\
+ Mixtral     & 5 & 0.940 & 0.941 & 0.936 & 0.942 & 0.946 & 0.924 \\
+ Moonshot    & 6 & 0.884 & 0.935 & 0.780 & 0.898 & 0.942 & 0.810 \\
+ GPT-4o-mini & 7 & 0.833 & 0.906 & 0.735 & 0.850 & 0.915 & 0.764 \\
+ Llama-3.1   & 8 & 0.795 & 0.888 & 0.702 & 0.812 & 0.895 & 0.729 \\
\bottomrule
\end{tabular}}
\end{table}

{
\mypara{Scaling the number of classes}
To probe larger class sets and the stability of the closed-form solver, we merge the source classes of RAID~\cite{DBLP:conf/acl/DuganHTZLXIC24} with those of our original benchmarks, remove overlapping models, and obtain 18 chronologically ordered source classes, with Human treated as one source class. The encoder is fine-tuned on the initial 6 classes and then frozen, after which 12 additional classes are added in sequence. We compare \method with SimpleCIL under the same frozen encoder, where SimpleCIL keeps a nearest-class-mean head and \method solves ridge regression on the calibrated and lifted representations, so the two already differ at the initial 6 classes.
As reported in \Cref{tab:class_scaling}, \method decreases from 0.965 to 0.779 full-F1 as the class count grows from 6 to 18, whereas SimpleCIL drops from 0.943 to 0.290. All closed-form solves complete successfully at the reported checkpoints, so the growing class count does not by itself cause numerical failure of the ridge solver. Increasing the random-feature dimension $d_\phi$ from 4096 to 8192 yields only a small further gain at the largest scale (full-F1 from 0.779 to 0.789 at 18 classes), indicating that the default $d_\phi=4096$ already provides sufficient capacity over this range. Random-feature capacity thus becomes more useful as classes accumulate, yet does not need to grow proportionally with the class count over this range.
}

\begin{table}[h!]
\centering

\caption{Scaling the number of source classes on RAID merged with our benchmarks. \method and SimpleCIL use the same frozen encoder. Full/New denote full macro-F1 and latest-class F1.}
\label{tab:class_scaling}
\small
\begin{tabular}{lcccc}
\toprule
\multirow{2}{*}{Classes} & \multicolumn{2}{c}{\method} & \multicolumn{2}{c}{SimpleCIL} \\
\cmidrule(lr){2-3}\cmidrule(lr){4-5}
 & Full & New & Full & New \\
\midrule
6  & 0.965 & --    & 0.943 & --    \\
10 & 0.913 & 0.902 & 0.585 & 0.299 \\
14 & 0.840 & 0.804 & 0.372 & 0.154 \\
18 & 0.779 & 0.742 & 0.290 & 0.144 \\
\bottomrule
\end{tabular}
\end{table}

\section{Independent Benchmark, Adversarial Robustness, and Error Analysis}
\label{app:extra}

\mypara{Validation on an independent benchmark}
To test \method on data built by a different group, we replicate the P5 protocol on RAID~\cite{DBLP:conf/acl/DuganHTZLXIC24} using the same DeBERTa-v3-base recipe and baseline implementations, with GPT-4 introduced as the new class at the final step. As shown in \Cref{tab:raid_p5}, \method attains the best Full, Old, and New F1 among all methods. Extending the setup to all 12 RAID source classes, with 6 initial and 6 incremental classes, \method reaches 0.857 full-F1 against 0.740 for the strongest baseline, iCaRL. The advantage of \method therefore transfers to an independently sourced benchmark and holds under a larger class set.

\begin{table}[h!]
\centering

\caption{Attribution performance on the independently constructed RAID benchmark under the P5 protocol, with GPT-4 as the new class. All methods use the same DeBERTa-v3-base backbone.}
\label{tab:raid_p5}
\small
\begin{tabular}{lccc}
\toprule
Method & Full-F1 & Old-F1 & New-F1 \\
\midrule
\method (Ours) & \textbf{0.977} & \textbf{0.976} & \textbf{0.983} \\
EASE      & 0.944 & 0.953 & 0.902 \\
BiC       & 0.942 & 0.955 & 0.873 \\
LwF       & 0.939 & 0.955 & 0.858 \\
iCaRL     & 0.939 & 0.954 & 0.861 \\
PASS      & 0.917 & 0.938 & 0.808 \\
SimpleCIL & 0.852 & 0.901 & 0.611 \\
\bottomrule
\end{tabular}
\end{table}

\mypara{Adversarial robustness}
We evaluate \method under a DIPPER paraphrasing attack~\cite{krishna2023paraphrasing} in the STEM / DeBERTa-v3-base / P5 setting with GPT-4o-mini as the new class. Attribution accuracy on machine-generated texts drops from 0.896 on clean inputs to 0.264 after rewriting, and SimpleCIL on the same frozen features reaches 0.144, suggesting that paraphrasing substantially weakens the generator-specific cues exploited by both classifiers. Because \method leaves the incremental update untouched, robustness can be injected through the base stage: adversarially training the DeBERTa-v3-base encoder before freezing raises the attacked-text accuracy from 0.264 to 0.604 under the same protocol. For reproducibility, we attack with DIPPER (the \texttt{dipper-paraphraser-xxl} model) in bfloat16, paraphrasing each text sentence by sentence with the preceding output as context and sampling with top-$p=0.75$ and lexical and order diversity both set to 60. Only the machine-generated test texts are rewritten (100 per generator, 500 in total) and the Human texts are never attacked; the rewritten texts are re-encoded by the same frozen encoder, and on this machine-only subset we report accuracy, the fraction still attributed to the true generator, since per-class F1 is ill-defined once the Human class is absent. For the robust variant, we augment only the base-stage initial-class data with paraphrases (about a $12\%$ augmentation ratio) and leave the incremental protocol unchanged, so no new-class paraphrase enters encoder training.

\begin{figure*}[t]
    \centering
    \includegraphics[width=0.80\textwidth]{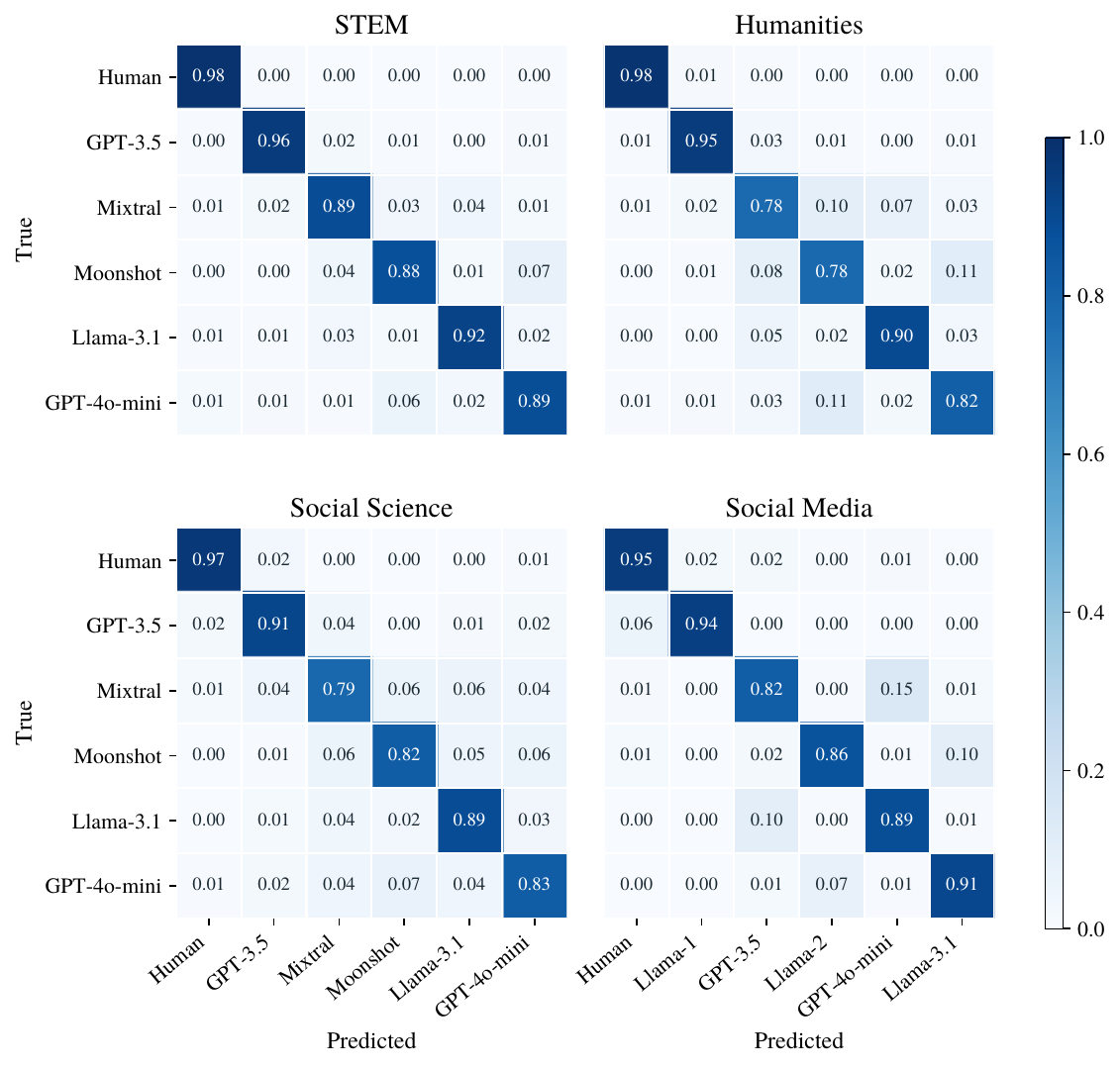}
    \caption{Row-normalized confusion matrices of the final P5 DeBERTa-v3-base \method model across the four evaluation domains, with GPT-4o-mini as the newly introduced class. Human texts are rarely misassigned to machine classes, and the residual errors concentrate between generators of the same family or writing style.}
    \label{fig:confusion}
\end{figure*}

\mypara{Misattribution patterns}
Using the final P5 DeBERTa-v3-base predictions with GPT-4o-mini as the new class, we compute row-normalized confusion matrices for each domain (\Cref{fig:confusion}). Human texts are rarely assigned to machine classes, with Human recall between 0.95 and 0.98 across domains. The remaining errors concentrate between generators of the same family or writing style: in the AIGTBench social-media domain, the largest confusions occur between GPT-3.5 and GPT-4o-mini (0.148 and 0.100 in the two directions) and between Llama-2 and Llama-3.1 (0.099 and 0.071). The machine-source errors of \method are thus structured around a few closely related generators rather than spread uniformly.

\section{Artifact Licenses, Intended Use, and Data Safety}
We use two publicly released benchmark datasets, MGT-Academic~\cite{DBLP:conf/kdd/LiuZL0ZWGT00025} and AIGTBench~\cite{DBLP:conf/acl/0001Z0ZL0Z025}, and do not collect any new user data.
MGT-Academic is released under the MIT license, and AIGTBench is released under the Apache 2.0 license.
Both datasets are released for research on machine-generated or AI-generated text detection/attribution, and our use is consistent with their intended research purposes.
We cite the original datasets and papers, and we do not redistribute the raw datasets as part of this work.

For privacy and content safety, we use the processed benchmark versions released by the original authors.
These datasets have already undergone preprocessing and moderation by their respective creators.
Our experiments only use text content and class labels for aggregate model training and evaluation, and we do not analyze, expose, or redistribute any individual-level information.

\section{Software Implementations and Package Settings}
All experiments were conducted in a Linux environment with kernel version 5.4.0-174-generic and Python 3.12.2.
We used PyTorch 2.10.0+cu128 with CUDA 12.8 on an NVIDIA L20 GPU with driver version 550.54.15.
Transformer encoders and tokenizers were implemented with HuggingFace Transformers 4.57.6, using DeBERTa-v3-base and RoBERTa-base as the main backbones.
Encoder fine-tuning followed a unified configuration with a maximum sequence length of 256, batch size 16, 3 training epochs, AdamW optimization, weight decay, and gradient clipping.

Traditional machine learning classifiers and evaluation metrics were implemented with scikit-learn 1.7.2.
Numerical computation relied on NumPy 2.0.1 and SciPy 1.15.2, while result aggregation and visualization used Pandas 2.2.3 and Matplotlib 3.9.1.
For \method, we fixed the main hyperparameters across all primary experiments: calibration exponent $\delta=0.5$, shrinkage coefficient $\alpha=0.05$, random feature dimension $d_\phi=4096$, ridge regularization coefficient $\lambda=1.0$, and class-balanced ridge coefficient $\beta=1.0$. Each experiment was run three times, and we report the average results.

\section{Computational Experiments}

We report the computational cost of different update strategies on a single NVIDIA L20 GPU.
Under the P5 protocol with DeBERTa-v3-base, \method updates the classifier by recomputing sufficient statistics and solving a closed-form ridge regression problem, which only takes a few seconds for the newly added generator.
In contrast, the joint-from-scratch baseline retrains an 184M-parameter DeBERTa-v3-base model on all old and new training texts, involving about 94k samples, and takes approximately 50 minutes on a single GPU.
The initial five-class encoder training is shared across methods and takes 1,738 seconds.

These results show that \method achieves much lower incremental update cost than full retraining, while avoiding access to old raw texts during the update stage.
This efficiency advantage is central to our lifelong MGT attribution setting, where attribution systems must incorporate newly emerging generators without repeatedly retraining the entire model.

\section{Use of AI Assistants}
We used AI assistants only for language polishing, grammar checking, and improving the readability of the manuscript.
All research ideas, experimental designs, implementation decisions, experimental results, analyses, and final claims were developed, checked, and approved by the authors.
The AI assistants were not used to independently generate, validate, or interpret experimental results.

\begin{table*}[t]
    \centering
    \caption{New-class F1 across Domains and Models (final continual step). $\ast$ denotes methods using data replay. Darker color indicates better performance. Bold values are the highest among continual-learning methods of the RoBERTa and DeBERTa blocks in each row. The ``Ori.''\ column is marked ``\textemdash'' because no new class has been introduced at S0.}
    \label{tab:newclass_full}
    \resizebox{\textwidth}{!}{
    \begin{tabular}{llcccccccccccccccc}
    \toprule
    \multirow{3}{*}{\textbf{Domain}} & \multirow{3}{*}{\textbf{New Model}} & \multicolumn{8}{c}{\textbf{RoBERTa-base}} & \multicolumn{8}{c}{\textbf{DeBERTa-base}} \\
    \cmidrule(lr){3-10} \cmidrule(lr){11-18}
    & & Ori. & LwF$\ast$ & iCaRL$\ast$ & BiC$\ast$ & EASE$\ast$ & PASS & SimpleCIL & \method (Ours) & Ori. & LwF$\ast$ & iCaRL$\ast$ & BiC$\ast$ & EASE$\ast$ & PASS & SimpleCIL & \method (Ours) \\
    \midrule
    
    \multirow{6}{*}{Social Science}
    & GPT-3.5     
    & — & \mgtScore{0.709} & \mgtScore{0.730} & \mgtScore{0.701} & \mgtScore{0.741} & \mgtScore{0.591} & \mgtScore{0.435} & \mgtBest{0.804}
    & — & \mgtScore{0.678} & \mgtScore{0.648} & \mgtScore{0.634} & \mgtScore{0.446} & \mgtScore{0.611} & \mgtScore{0.478} & \mgtBest{0.785} \\
    
    & Mixtral     
    & — & \mgtScore{0.667} & \mgtScore{0.684} & \mgtScore{0.677} & \mgtScore{0.665} & \mgtScore{0.621} & \mgtScore{0.356} & \mgtBest{0.760}
    & — & \mgtScore{0.632} & \mgtScore{0.652} & \mgtScore{0.630} & \mgtScore{0.420} & \mgtScore{0.592} & \mgtScore{0.399} & \mgtBest{0.750} \\
    
    & Moonshot     
    & — & \mgtScore{0.675} & \mgtScore{0.679} & \mgtScore{0.655} & \mgtScore{0.557} & \mgtScore{0.590} & \mgtScore{0.477} & \mgtBest{0.738}
    & — & \mgtScore{0.644} & \mgtScore{0.658} & \mgtScore{0.646} & \mgtScore{0.473} & \mgtScore{0.590} & \mgtScore{0.430} & \mgtBest{0.740} \\
    
    & Llama-3.1     
    & — & \mgtScore{0.592} & \mgtScore{0.591} & \mgtScore{0.572} & \mgtScore{0.618} & \mgtScore{0.560} & \mgtScore{0.332} & \mgtBest{0.716}
    & — & \mgtScore{0.457} & \mgtScore{0.555} & \mgtScore{0.486} & \mgtScore{0.395} & \mgtScore{0.541} & \mgtScore{0.347} & \mgtBest{0.742} \\
    
    & GPT-4o-mini     
    & — & \mgtScore{0.676} & \mgtScore{0.662} & \mgtScore{0.721} & \mgtScore{0.696} & \mgtScore{0.591} & \mgtScore{0.546} & \mgtBest{0.808}
    & — & \mgtScore{0.713} & \mgtScore{0.718} & \mgtScore{0.710} & \mgtScore{0.705} & \mgtScore{0.646} & \mgtScore{0.481} & \mgtBest{0.830} \\
    
    \cmidrule(lr){2-18}
    & Average     
    & — & \mgtScore{0.664} & \mgtScore{0.669} & \mgtScore{0.665} & \mgtScore{0.655} & \mgtScore{0.591} & \mgtScore{0.429} & \mgtBest{0.765}
    & — & \mgtScore{0.625} & \mgtScore{0.646} & \mgtScore{0.621} & \mgtScore{0.488} & \mgtScore{0.596} & \mgtScore{0.427} & \mgtBest{0.769} \\
    \midrule
    
    \multirow{6}{*}{STEM}
    & GPT-3.5     
    & — & \mgtScore{0.767} & \mgtScore{0.761} & \mgtScore{0.751} & \mgtScore{0.737} & \mgtScore{0.610} & \mgtScore{0.460} & \mgtBest{0.857}
    & — & \mgtScore{0.690} & \mgtScore{0.707} & \mgtScore{0.677} & \mgtScore{0.665} & \mgtScore{0.615} & \mgtScore{0.459} & \mgtBest{0.845} \\
    
    & Mixtral     
    & — & \mgtScore{0.724} & \mgtScore{0.728} & \mgtScore{0.693} & \mgtScore{0.719} & \mgtScore{0.637} & \mgtScore{0.424} & \mgtBest{0.804}
    & — & \mgtScore{0.722} & \mgtScore{0.708} & \mgtScore{0.686} & \mgtScore{0.711} & \mgtScore{0.614} & \mgtScore{0.476} & \mgtBest{0.825} \\
    
    & Moonshot     
    & — & \mgtScore{0.751} & \mgtScore{0.710} & \mgtScore{0.747} & \mgtScore{0.746} & \mgtScore{0.663} & \mgtScore{0.512} & \mgtBest{0.823}
    & — & \mgtScore{0.766} & \mgtScore{0.771} & \mgtScore{0.760} & \mgtScore{0.768} & \mgtScore{0.660} & \mgtScore{0.609} & \mgtBest{0.839} \\
    
    & Llama-3.1     
    & — & \mgtScore{0.666} & \mgtScore{0.582} & \mgtScore{0.629} & \mgtScore{0.589} & \mgtScore{0.612} & \mgtScore{0.432} & \mgtBest{0.787}
    & — & \mgtScore{0.677} & \mgtScore{0.669} & \mgtScore{0.625} & \mgtScore{0.673} & \mgtScore{0.616} & \mgtScore{0.519} & \mgtBest{0.821} \\
    
    & GPT-4o-mini     
    & — & \mgtScore{0.798} & \mgtScore{0.784} & \mgtScore{0.800} & \mgtScore{0.644} & \mgtScore{0.661} & \mgtScore{0.567} & \mgtBest{0.871}
    & — & \mgtScore{0.731} & \mgtScore{0.756} & \mgtScore{0.792} & \mgtScore{0.813} & \mgtScore{0.681} & \mgtScore{0.556} & \mgtBest{0.881} \\
    
    \cmidrule(lr){2-18}
    & Average     
    & — & \mgtScore{0.741} & \mgtScore{0.713} & \mgtScore{0.724} & \mgtScore{0.687} & \mgtScore{0.637} & \mgtScore{0.479} & \mgtBest{0.829}
    & — & \mgtScore{0.717} & \mgtScore{0.722} & \mgtScore{0.708} & \mgtScore{0.726} & \mgtScore{0.637} & \mgtScore{0.524} & \mgtBest{0.842} \\
    \midrule
    
    \multirow{6}{*}{Humanities}
    & GPT-3.5     
    & — & \mgtScore{0.806} & \mgtScore{0.805} & \mgtScore{0.794} & \mgtScore{0.815} & \mgtScore{0.684} & \mgtScore{0.496} & \mgtBest{0.867}
    & — & \mgtScore{0.713} & \mgtScore{0.756} & \mgtScore{0.741} & \mgtScore{0.738} & \mgtScore{0.652} & \mgtScore{0.525} & \mgtBest{0.858} \\
    
    & Mixtral     
    & — & \mgtScore{0.645} & \mgtScore{0.628} & \mgtScore{0.610} & \mgtScore{0.632} & \mgtScore{0.612} & \mgtScore{0.420} & \mgtBest{0.696}
    & — & \mgtScore{0.615} & \mgtScore{0.604} & \mgtScore{0.575} & \mgtScore{0.558} & \mgtScore{0.589} & \mgtScore{0.423} & \mgtBest{0.690} \\
    
    & Moonshot     
    & — & \mgtScore{0.661} & \mgtScore{0.665} & \mgtScore{0.635} & \mgtScore{0.636} & \mgtScore{0.589} & \mgtScore{0.542} & \mgtBest{0.709}
    & — & \mgtScore{0.644} & \mgtScore{0.643} & \mgtScore{0.634} & \mgtScore{0.630} & \mgtScore{0.596} & \mgtScore{0.557} & \mgtBest{0.723} \\
    
    & Llama-3.1     
    & — & \mgtScore{0.525} & \mgtScore{0.607} & \mgtScore{0.518} & \mgtScore{0.543} & \mgtScore{0.587} & \mgtScore{0.416} & \mgtBest{0.693}
    & — & \mgtScore{0.567} & \mgtScore{0.509} & \mgtScore{0.530} & \mgtScore{0.515} & \mgtScore{0.560} & \mgtScore{0.441} & \mgtBest{0.735} \\
    
    & GPT-4o-mini     
    & — & \mgtScore{0.733} & \mgtScore{0.741} & \mgtScore{0.752} & \mgtScore{0.743} & \mgtScore{0.652} & \mgtScore{0.480} & \mgtBest{0.827}
    & — & \mgtScore{0.686} & \mgtScore{0.632} & \mgtScore{0.693} & \mgtScore{0.682} & \mgtScore{0.667} & \mgtScore{0.364} & \mgtBest{0.823} \\
    
    \cmidrule(lr){2-18}
    & Average     
    & — & \mgtScore{0.674} & \mgtScore{0.689} & \mgtScore{0.662} & \mgtScore{0.674} & \mgtScore{0.625} & \mgtScore{0.471} & \mgtBest{0.758}
    & — & \mgtScore{0.645} & \mgtScore{0.629} & \mgtScore{0.635} & \mgtScore{0.625} & \mgtScore{0.612} & \mgtScore{0.462} & \mgtBest{0.766} \\
    \midrule
    
    \multirow{6}{*}{AIGTBench}
    & GPT-3.5     
    & — & \mgtScore{0.678} & \mgtScore{0.646} & \mgtScore{0.674} & \mgtScore{0.643} & \mgtScore{0.667} & \mgtScore{0.363} & \mgtBest{0.745}
    & — & \mgtScore{0.720} & \mgtScore{0.708} & \mgtScore{0.660} & \mgtScore{0.705} & \mgtScore{0.676} & \mgtScore{0.468} & \mgtBest{0.755} \\
    
    & GPT-4o-mini     
    & — & \mgtScore{0.701} & \mgtScore{0.558} & \mgtScore{0.678} & \mgtScore{0.580} & \mgtScore{0.732} & \mgtScore{0.507} & \mgtBest{0.789}
    & — & \mgtScore{0.682} & \mgtScore{0.696} & \mgtScore{0.598} & \mgtScore{0.685} & \mgtScore{0.690} & \mgtScore{0.494} & \mgtBest{0.775} \\
    
    & Llama-1     
    & — & \mgtScore{0.903} & \mgtScore{0.851} & \mgtScore{0.917} & \mgtScore{0.922} & \mgtScore{0.810} & \mgtScore{0.845} & \mgtBest{0.960}
    & — & \mgtScore{0.877} & \mgtScore{0.873} & \mgtScore{0.878} & \mgtScore{0.882} & \mgtScore{0.772} & \mgtScore{0.625} & \mgtBest{0.943} \\
    
    & Llama-2     
    & — & \mgtScore{0.751} & \mgtScore{0.708} & \mgtScore{0.748} & \mgtScore{0.724} & \mgtScore{0.650} & \mgtScore{0.409} & \mgtBest{0.867}
    & — & \mgtScore{0.730} & \mgtScore{0.769} & \mgtScore{0.746} & \mgtScore{0.763} & \mgtScore{0.638} & \mgtScore{0.574} & \mgtBest{0.862} \\
    
    & Llama-3.1     
    & — & \mgtScore{0.780} & \mgtScore{0.758} & \mgtScore{0.815} & \mgtScore{0.794} & \mgtScore{0.735} & \mgtScore{0.487} & \mgtBest{0.907}
    & — & \mgtScore{0.743} & \mgtScore{0.678} & \mgtScore{0.715} & \mgtScore{0.756} & \mgtScore{0.663} & \mgtScore{0.557} & \mgtBest{0.894} \\
    
    \cmidrule(lr){2-18}
    & Average     
    & — & \mgtScore{0.763} & \mgtScore{0.704} & \mgtScore{0.766} & \mgtScore{0.733} & \mgtScore{0.719} & \mgtScore{0.522} & \mgtBest{0.853}
    & — & \mgtScore{0.750} & \mgtScore{0.745} & \mgtScore{0.719} & \mgtScore{0.758} & \mgtScore{0.688} & \mgtScore{0.543} & \mgtBest{0.846} \\
    \midrule
    
    \multicolumn{2}{l}{Overall Average}
    & — & \mgtScore{0.710} & \mgtScore{0.694} & \mgtScore{0.704} & \mgtScore{0.687} & \mgtScore{0.643} & \mgtScore{0.475} & \mgtBest{0.801}
    & — & \mgtScore{0.684} & \mgtScore{0.685} & \mgtScore{0.671} & \mgtScore{0.649} & \mgtScore{0.633} & \mgtScore{0.489} & \mgtBest{0.806} \\
    \bottomrule
    \end{tabular}
    }
    \end{table*}

\begin{table*}[t]
    \centering
    \caption{Old-class F1 across domains and models (final incremental step, initial-class subset). $\ast$ denotes methods using data replay. Darker color indicates better performance. Bold values indicate the best result among continual-learning methods within each RoBERTa-base and DeBERTa-base block. The ``Ori.'' column reports the S0 macro-F1 on the initial classes, which is identical to full-F1 because all classes are old at S0.}
    \label{tab:oldclass_full}
    \resizebox{\textwidth}{!}{
    \begin{tabular}{llcccccccccccccccc}
    \toprule
    \multirow{3}{*}{\textbf{Domain}} & \multirow{3}{*}{\textbf{New Model}} & \multicolumn{8}{c}{\textbf{RoBERTa-base}} & \multicolumn{8}{c}{\textbf{DeBERTa-base}} \\
    \cmidrule(lr){3-10} \cmidrule(lr){11-18}
    & & Ori. & LwF$\ast$ & iCaRL$\ast$ & BiC$\ast$ & EASE$\ast$ & PASS & SimpleCIL & \method (Ours) & Ori. & LwF$\ast$ & iCaRL$\ast$ & BiC$\ast$ & EASE$\ast$ & PASS & SimpleCIL & \method (Ours) \\
    \midrule
    
    \multirow{6}{*}{Social Science}
    & GPT-3.5     
    & 0.906 & \mgtScore{0.858} & \mgtScore{0.868} & \mgtScore{0.861} & \mgtScore{0.864} & \mgtScore{0.808} & \mgtScore{0.836} & \mgtBest{0.877}
    & 0.891 & \mgtScore{0.847} & \mgtScore{0.845} & \mgtScore{0.835} & \mgtScore{0.814} & \mgtScore{0.814} & \mgtScore{0.822} & \mgtBest{0.864} \\
    
    & Mixtral     
    & 0.938 & \mgtScore{0.881} & \mgtScore{0.868} & \mgtScore{0.886} & \mgtScore{0.881} & \mgtScore{0.837} & \mgtScore{0.854} & \mgtBest{0.905}
    & 0.938 & \mgtScore{0.883} & \mgtScore{0.866} & \mgtScore{0.873} & \mgtScore{0.826} & \mgtScore{0.824} & \mgtScore{0.848} & \mgtBest{0.903} \\
    
    & Moonshot     
    & 0.931 & \mgtScore{0.881} & \mgtScore{0.875} & \mgtScore{0.877} & \mgtScore{0.868} & \mgtScore{0.818} & \mgtScore{0.855} & \mgtBest{0.893}
    & 0.922 & \mgtScore{0.873} & \mgtScore{0.863} & \mgtScore{0.864} & \mgtScore{0.850} & \mgtScore{0.822} & \mgtScore{0.847} & \mgtBest{0.886} \\
    
    & Llama-3.1     
    & 0.920 & \mgtScore{0.865} & \mgtScore{0.849} & \mgtScore{0.863} & \mgtScore{0.850} & \mgtScore{0.801} & \mgtScore{0.835} & \mgtBest{0.882}
    & 0.912 & \mgtScore{0.850} & \mgtScore{0.837} & \mgtScore{0.844} & \mgtScore{0.824} & \mgtScore{0.792} & \mgtScore{0.818} & \mgtBest{0.875} \\
    
    & GPT-4o-mini     
    & 0.905 & \mgtScore{0.860} & \mgtScore{0.840} & \mgtScore{0.862} & \mgtScore{0.832} & \mgtScore{0.804} & \mgtScore{0.839} & \mgtBest{0.879}
    & 0.898 & \mgtScore{0.859} & \mgtScore{0.844} & \mgtScore{0.846} & \mgtScore{0.854} & \mgtScore{0.825} & \mgtScore{0.822} & \mgtBest{0.874} \\
    
    \cmidrule(lr){2-18}
    & Average     
    & 0.920 & \mgtScore{0.869} & \mgtScore{0.860} & \mgtScore{0.870} & \mgtScore{0.859} & \mgtScore{0.814} & \mgtScore{0.844} & \mgtBest{0.887}
    & 0.912 & \mgtScore{0.862} & \mgtScore{0.851} & \mgtScore{0.852} & \mgtScore{0.834} & \mgtScore{0.815} & \mgtScore{0.831} & \mgtBest{0.881} \\
    \midrule
    
    \multirow{6}{*}{STEM}
    & GPT-3.5     
    & 0.946 & \mgtScore{0.905} & \mgtScore{0.903} & \mgtScore{0.909} & \mgtScore{0.892} & \mgtScore{0.841} & \mgtScore{0.874} & \mgtBest{0.925}
    & 0.942 & \mgtScore{0.878} & \mgtScore{0.879} & \mgtScore{0.891} & \mgtScore{0.883} & \mgtScore{0.847} & \mgtScore{0.869} & \mgtBest{0.919} \\
    
    & Mixtral     
    & 0.964 & \mgtScore{0.920} & \mgtScore{0.914} & \mgtScore{0.914} & \mgtScore{0.904} & \mgtScore{0.869} & \mgtScore{0.886} & \mgtBest{0.933}
    & 0.966 & \mgtScore{0.911} & \mgtScore{0.911} & \mgtScore{0.912} & \mgtScore{0.898} & \mgtScore{0.853} & \mgtScore{0.891} & \mgtBest{0.938} \\
    
    & Moonshot     
    & 0.959 & \mgtScore{0.924} & \mgtScore{0.914} & \mgtScore{0.920} & \mgtScore{0.912} & \mgtScore{0.877} & \mgtScore{0.893} & \mgtBest{0.934}
    & 0.958 & \mgtScore{0.918} & \mgtScore{0.908} & \mgtScore{0.914} & \mgtScore{0.916} & \mgtScore{0.869} & \mgtScore{0.902} & \mgtBest{0.932} \\
    
    & Llama-3.1     
    & 0.959 & \mgtScore{0.889} & \mgtScore{0.896} & \mgtScore{0.895} & \mgtScore{0.790} & \mgtScore{0.848} & \mgtScore{0.875} & \mgtBest{0.925}
    & 0.956 & \mgtScore{0.896} & \mgtScore{0.889} & \mgtScore{0.897} & \mgtScore{0.893} & \mgtScore{0.847} & \mgtScore{0.883} & \mgtBest{0.928} \\
    
    & GPT-4o-mini     
    & 0.942 & \mgtScore{0.906} & \mgtScore{0.901} & \mgtScore{0.902} & \mgtScore{0.808} & \mgtScore{0.859} & \mgtScore{0.879} & \mgtBest{0.921}
    & 0.945 & \mgtScore{0.879} & \mgtScore{0.883} & \mgtScore{0.909} & \mgtScore{0.903} & \mgtScore{0.868} & \mgtScore{0.886} & \mgtBest{0.927} \\
    
    \cmidrule(lr){2-18}
    & Average     
    & 0.954 & \mgtScore{0.909} & \mgtScore{0.906} & \mgtScore{0.908} & \mgtScore{0.861} & \mgtScore{0.859} & \mgtScore{0.881} & \mgtBest{0.928}
    & 0.953 & \mgtScore{0.896} & \mgtScore{0.894} & \mgtScore{0.905} & \mgtScore{0.899} & \mgtScore{0.857} & \mgtScore{0.886} & \mgtBest{0.929} \\
    \midrule
    
    \multirow{6}{*}{Humanities}
    & GPT-3.5     
    & 0.903 & \mgtScore{0.864} & \mgtScore{0.866} & \mgtScore{0.838} & \mgtScore{0.872} & \mgtScore{0.842} & \mgtScore{0.837} & \mgtBest{0.884}
    & 0.898 & \mgtScore{0.849} & \mgtScore{0.838} & \mgtScore{0.853} & \mgtScore{0.844} & \mgtScore{0.814} & \mgtScore{0.816} & \mgtBest{0.876} \\
    
    & Mixtral     
    & 0.954 & \mgtScore{0.898} & \mgtScore{0.899} & \mgtScore{0.899} & \mgtScore{0.893} & \mgtScore{0.857} & \mgtScore{0.882} & \mgtBest{0.912}
    & 0.952 & \mgtScore{0.874} & \mgtScore{0.872} & \mgtScore{0.891} & \mgtScore{0.887} & \mgtScore{0.846} & \mgtScore{0.864} & \mgtBest{0.906} \\
    
    & Moonshot     
    & 0.936 & \mgtScore{0.889} & \mgtScore{0.874} & \mgtScore{0.885} & \mgtScore{0.885} & \mgtScore{0.817} & \mgtScore{0.860} & \mgtBest{0.893}
    & 0.945 & \mgtScore{0.861} & \mgtScore{0.866} & \mgtScore{0.882} & \mgtScore{0.862} & \mgtScore{0.828} & \mgtScore{0.866} & \mgtBest{0.902} \\
    
    & Llama-3.1     
    & 0.928 & \mgtScore{0.866} & \mgtScore{0.839} & \mgtScore{0.863} & \mgtScore{0.863} & \mgtScore{0.808} & \mgtScore{0.842} & \mgtBest{0.882}
    & 0.922 & \mgtScore{0.727} & \mgtScore{0.843} & \mgtScore{0.848} & \mgtScore{0.844} & \mgtScore{0.774} & \mgtScore{0.834} & \mgtBest{0.880} \\
    
    & GPT-4o-mini     
    & 0.903 & \mgtScore{0.865} & \mgtScore{0.855} & \mgtScore{0.867} & \mgtScore{0.857} & \mgtScore{0.828} & \mgtScore{0.829} & \mgtBest{0.876}
    & 0.905 & \mgtScore{0.859} & \mgtScore{0.769} & \mgtScore{0.857} & \mgtScore{0.831} & \mgtScore{0.833} & \mgtScore{0.804} & \mgtBest{0.877} \\
    
    \cmidrule(lr){2-18}
    & Average     
    & 0.925 & \mgtScore{0.876} & \mgtScore{0.867} & \mgtScore{0.870} & \mgtScore{0.874} & \mgtScore{0.831} & \mgtScore{0.850} & \mgtBest{0.889}
    & 0.924 & \mgtScore{0.834} & \mgtScore{0.838} & \mgtScore{0.866} & \mgtScore{0.853} & \mgtScore{0.819} & \mgtScore{0.837} & \mgtBest{0.888} \\
    \midrule
    
    \multirow{6}{*}{AIGTBench}
    & GPT-3.5     
    & 0.964 & \mgtScore{0.915} & \mgtScore{0.915} & \mgtScore{0.916} & \mgtScore{0.907} & \mgtScore{0.875} & \mgtScore{0.892} & \mgtBest{0.924}
    & 0.959 & \mgtScore{0.903} & \mgtScore{0.874} & \mgtScore{0.903} & \mgtScore{0.880} & \mgtScore{0.864} & \mgtScore{0.883} & \mgtBest{0.920} \\
    
    & GPT-4o-mini     
    & 0.956 & \mgtScore{0.847} & \mgtScore{0.888} & \mgtScore{0.903} & \mgtScore{0.844} & \mgtScore{0.879} & \mgtScore{0.882} & \mgtBest{0.918}
    & 0.954 & \mgtScore{0.821} & \mgtScore{0.862} & \mgtScore{0.881} & \mgtScore{0.825} & \mgtScore{0.840} & \mgtScore{0.864} & \mgtBest{0.912} \\
    
    & Llama-1     
    & 0.913 & \mgtScore{0.890} & \mgtScore{0.870} & \mgtScore{0.893} & \mgtScore{0.891} & \mgtScore{0.859} & \mgtScore{0.884} & \mgtBest{0.906}
    & 0.916 & \mgtScore{0.882} & \mgtScore{0.869} & \mgtScore{0.872} & \mgtScore{0.872} & \mgtScore{0.859} & \mgtScore{0.828} & \mgtBest{0.904} \\
    
    & Llama-2     
    & 0.918 & \mgtScore{0.877} & \mgtScore{0.838} & \mgtScore{0.876} & \mgtScore{0.860} & \mgtScore{0.824} & \mgtScore{0.848} & \mgtBest{0.897}
    & 0.919 & \mgtScore{0.867} & \mgtScore{0.854} & \mgtScore{0.878} & \mgtScore{0.861} & \mgtScore{0.825} & \mgtScore{0.859} & \mgtBest{0.899} \\
    
    & Llama-3.1     
    & 0.915 & \mgtScore{0.881} & \mgtScore{0.858} & \mgtScore{0.865} & \mgtScore{0.846} & \mgtScore{0.856} & \mgtScore{0.835} & \mgtBest{0.897}
    & 0.913 & \mgtScore{0.860} & \mgtScore{0.779} & \mgtScore{0.865} & \mgtScore{0.853} & \mgtScore{0.823} & \mgtScore{0.844} & \mgtBest{0.895} \\
    
    \cmidrule(lr){2-18}
    & Average     
    & 0.933 & \mgtScore{0.882} & \mgtScore{0.874} & \mgtScore{0.890} & \mgtScore{0.869} & \mgtScore{0.859} & \mgtScore{0.868} & \mgtBest{0.908}
    & 0.932 & \mgtScore{0.867} & \mgtScore{0.848} & \mgtScore{0.880} & \mgtScore{0.858} & \mgtScore{0.842} & \mgtScore{0.856} & \mgtBest{0.906} \\
    \midrule
    
    \multicolumn{2}{l}{Overall Average}
    & 0.933 & \mgtScore{0.884} & \mgtScore{0.877} & \mgtScore{0.885} & \mgtScore{0.866} & \mgtScore{0.840} & \mgtScore{0.861} & \mgtBest{0.903}
    & 0.930 & \mgtScore{0.865} & \mgtScore{0.857} & \mgtScore{0.876} & \mgtScore{0.861} & \mgtScore{0.833} & \mgtScore{0.853} & \mgtBest{0.901} \\
    \bottomrule
    \end{tabular}
    }
\end{table*}

\end{document}